\documentclass[conference]{IEEEtran}
\IEEEoverridecommandlockouts

\usepackage{cite}
\usepackage{amsmath,amssymb,amsfonts,graphicx}
\usepackage{algorithm}
\usepackage{algpseudocode}
\usepackage{graphicx}
\usepackage{textcomp}
\usepackage[dvipsnames]{xcolor}
\usepackage{url}
\usepackage{booktabs, multirow} 
\usepackage[T1]{fontenc}
\usepackage{float}
\usepackage{pifont}

\def\BibTeX{{\rm B\kern-.05em{\sc i\kern-.025em b}\kern-.08em
    T\kern-.1667em\lower.7ex\hbox{E}\kern-.125emX}}
\begin{document}

\title{Temporal Window Smoothing of Exogenous Variables for Improved Time Series Prediction}



\author{
\IEEEauthorblockN{Mustafa Kamal\textsuperscript{1,*}\thanks{* Equal contribution},
Niyaz Bin Hashem\textsuperscript{2,*},
Robin Krambroeckers\textsuperscript{1},
Nabeel Mohammed\textsuperscript{2},
Shafin Rahman\textsuperscript{2}}
\IEEEauthorblockA{\textsuperscript{1}Artificial Intelligence Department, RobotBulls Labs, Geneva, Switzerland}
\IEEEauthorblockA{\textsuperscript{2}Department of Electrical and Computer Engineering, North South University, Dhaka, Bangladesh}
\IEEEauthorblockA{
\{mustafa.kamal, robin\}@robotbulls.com, \\
\{nabeel.mohammed, shafin.rahman\}@northsouth.edu, \\
niyaz.hashem@gmail.com}
}

\maketitle
\begin{abstract}
Although most transformer-based time series forecasting models primarily depend on endogenous inputs, recent state-of-the-art approaches have significantly improved performance by incorporating external information through exogenous inputs. However, these methods face challenges, such as redundancy when endogenous and exogenous inputs originate from the same source and limited ability to capture long-term dependencies due to fixed look-back windows. In this paper, we propose a method that whitens the exogenous input to reduce redundancy that may persist within the data based on global statistics. Additionally, our approach helps the exogenous input to be more aware of patterns and trends over extended periods. By introducing this refined, globally context-aware exogenous input to the endogenous input without increasing the lookback window length, our approach guides the model towards improved forecasting. Our approach achieves state-of-the-art performance in four benchmark datasets, consistently outperforming 11 baseline models. These results establish our method as a robust and effective alternative for using exogenous inputs in time series forecasting.
\end{abstract}


\section{Introduction}
Time series forecasting is a fundamental problem in time series analysis. It is widely used in weather prediction \cite{wu2023interpretable,zhang2023skilful}, energy management \cite{lago2021forecasting, weron2014electricity}, finance \cite{zeng2023financial}, and transportation systems \cite{lv2014traffic}. The primary problem it solves is the uncertainty in future trends and patterns, helping organizations plan and optimize resources effectively. Time series forecasting enables businesses to mitigate risks, manage supply chains, allocate resources efficiently, and improve overall operational performance by accurately predicting future values. In these applications, the inclusion of exogenous variables that influence the target variable has proven essential to improve forecast precision \cite{vagropoulos2016comparison, wang2024timexer,williams2001multivariate,das2023long}. By incorporating exogenous variables, the models gain a deeper understanding of the underlying factors that drive the time series, leading to more precise and reliable predictions. In this paper, we propose a temporal window smoothing technique for exogenous variables based on global statistics to further improve the forecasting performance without adding any additional learnable parameters.

Although exogenous variables improve prediction performance, they often introduce noise, irregularities, redundancy, and inconsistencies. Moreover, their awareness is typically limited to trends within a specific window, restricting their ability to capture broader, longer-term patterns.
\begin{figure}
    \centering
    \includegraphics[width=1\linewidth]{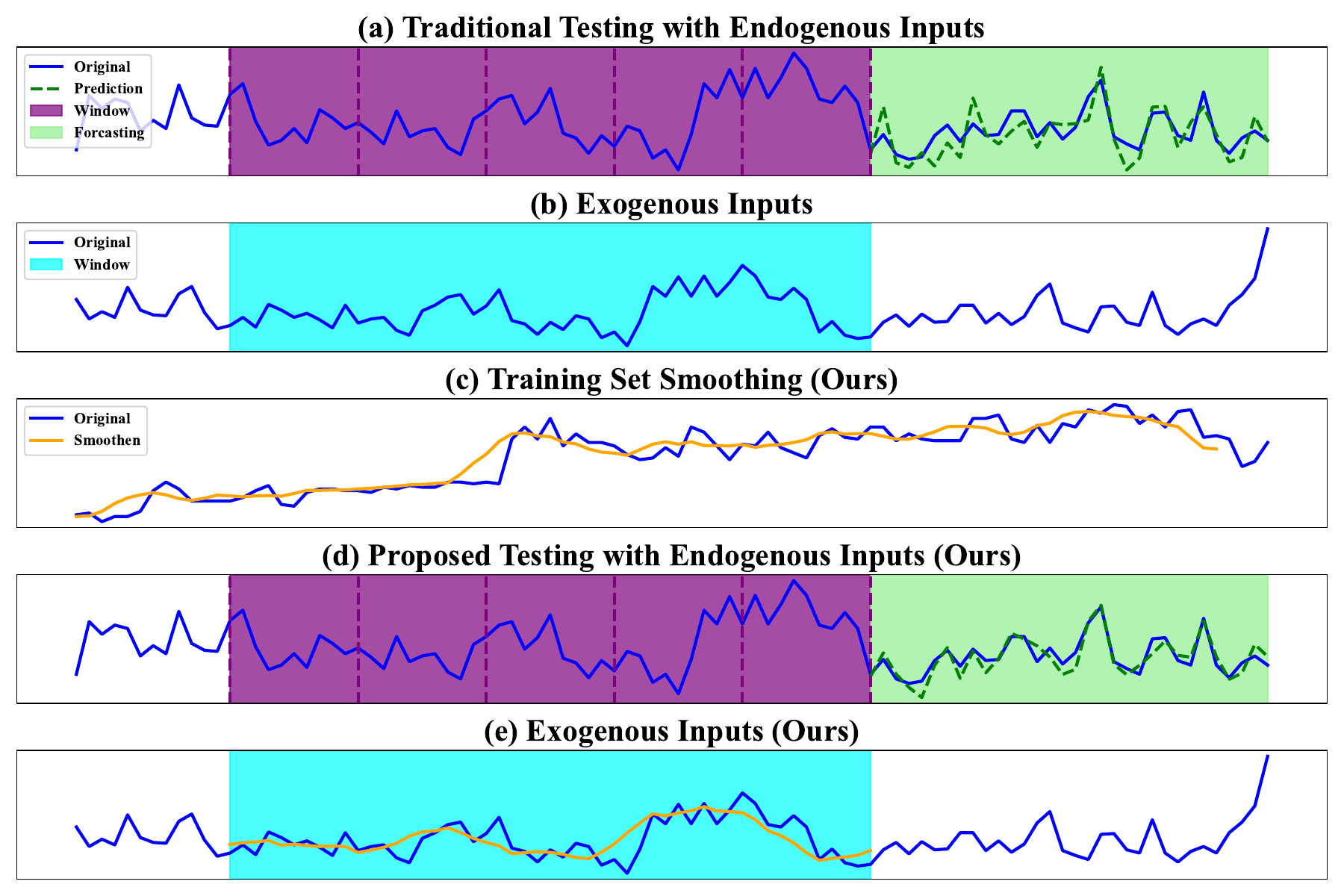}
    \caption{Visualization of our proposed approach for time series forecasting: (a) Traditional forecasting with endogenous inputs, where the window of endogenous series (highlighted in purple) contains multiple patches, and exogenous inputs (b), highlighted in blue, are incorporated with endogenous inputs to provide additional context. Predictions are denoted in green. In contrast, our approach first captures globally significant patterns from the entire training set (c). Based on this, we whiten each window of exogenous series (e) to make it more aware of these globally significant patterns before integrating it with the patches of endogenous series (d), leading to improved forecasting performance, as highlighted in green.} 
    \label{fig:residual}
\end{figure}

Addressing these issues is crucial for maximizing the benefits of exogenous variables, ensuring that they contribute to better model performance, more accurate predictions, and better generalization in real-world applications. Although some approaches in time series forecasting address challenges such as irregularities and missing values in data \cite{wang2024timexer,liu2023itransformer}, to the best of our knowledge, no notable work in transformer-based multivariate forecasting ensures that each window of the exogenous series is aware of global dependencies, which could help the model understand trends, seasonality, and patterns that span longer periods, while also addressing the challenges of redundancy in exogenous variables. Existing solutions often struggle to effectively handle exogenous variables, leading to suboptimal representations. There are two streams of work:
\textbf{\textit{(1)}} One common approach is to concatenate exogenous and endogenous variables before projecting them into a latent space \cite{das2023long}. \textbf{\textit{(2)}} Another method independently projects exogenous variables into a latent space before integrating them with endogenous variables through a token \cite{wang2024timexer}.  However, both scenarios risk introducing redundant information into the final forecast, potentially leading to suboptimal model performance. Furthermore, these approaches may fail to fully leverage exogenous variables to enhance the endogenous series' understanding of global patterns, such as long-term trends and dependencies.


To address these challenges, we introduce a Temporal Window Smoothing (TWS) block, designed to reduce redundant information from exogenous inputs while enhancing their awareness of global dependencies. The proposed approach projects each windowed time series onto a set of orthogonal basis vectors, derived globally from the entire training set. This exposes exogenous inputs to key patterns, such as long-term trends and seasonality, while filtering out globally insignificant values. Furthermore, a dynamic mechanism selects the optimal number of basis vectors, ensuring that significant patterns are captured without redundancy. Finally, the reconstruction mechanism generates a refined exogenous series as shown in Fig. \ref{fig:residual} by combining the selected basis vectors, preserving the original data dimensions while minimizing redundancy. This approach ensures that exogenous inputs provide a global context without propagating unnecessary or irrelevant information, enhancing the model’s ability to capture meaningful dependencies and improve forecasting accuracy. The proposed approach was evaluated on seven diverse datasets, each representing unique characteristics and challenges in time series forecasting, such as electricity, weather, and traffic. Our method consistently outperformed 11 baseline models in four of the seven benchmark datasets. The main contributions of this model are as follows: 
\begin{itemize}
    \item We propose a temporal window smoothing approach for exogenous variables for improved time series forecasting.
    \item We identify that the widespread use of exogenous variables potentially inserts redundant information within the forecasting network, which requires whitening to obtain the full benefit of the exogenous input.
    \item We expose exogenous inputs to globally significant patterns to improve their robustness. 
    \item We perform extensive experiments on benchmark datasets, including the ETT (comprising ETTm1, ETTm2, ETTh1, and ETTh2) \cite{zhou2021informer}, Weather \cite{zhou2021informer}, ECL \cite{li2019enhancing}, and Traffic \cite{Timesnet}, highlighting that our technique beats the latest results in most experimental settings.
\end{itemize}

\section{Related Work}


\noindent\textbf{Long-Sequence Time Series Forecasting with Transformers:} The remarkable success of Transformers in natural language processing \cite{devlin2018bert} and computer vision \cite{dosovitskiy2020image,liu2021swin} has driven growing interest in their application to time series analysis. Long-sequence time series forecasting (LSTF) requires models to have strong predictive capabilities, enabling them to effectively capture and utilize the relationships between inputs and outputs over extended time periods. Transformers have proven highly effective in capturing long-term temporal dependencies and modeling complex multivariate relationships, making them particularly well-suited for such tasks. Popular Transformer-based models can be divided into point-wise \cite{zhou2021informer,Autoformer,liu2021pyraformer,zhou2022fedformer,dong2023simmtm}, patch-wise \cite{PatchTST}, and variate-wise \cite{liu2023itransformer} approaches according to their representation granularity. Given the sequential nature of time series, early research focused on point-wise representations to model correlations between time points, leading to the development of efficient Transformer models to manage computational complexity. However, in the context of long-sequence time-series forecasting (LSTF), Transformers face significant challenges, including the quadratic time complexity of the attention mechanism and high memory usage, which make them less practical for direct application. Informer \cite{zhou2021informer} addresses these limitations by introducing an efficient transformer-based model that reduces the time complexity from quadratic to \( O(L \log L) \). Recognizing the limitations of point-wise representations in capturing local semantic information, PatchTST \cite{PatchTST} improves efficiency by aggregating time steps into subseries-level patches to improve locality, using channel-independent tokens for better representation of features, and reducing token length. In contrast, iTransformer\cite{liu2023itransformer} tackles issues such as weak multivariate correlation modeling and limited temporal representation by embedding entire time series of individual variates as tokens, allowing for better attention-based correlation modeling and global representation learning. Although the recent Transformer-based approach \cite{wang2024timexer} incorporates exogenous variables to jointly capture intra-endogenous temporal dependencies, they do not fully leverage the potential of exogenous or address the redundant information it may contain. In contrast, this paper proposes a novel technique that addresses these issues.

\noindent\textbf{Exogenous Variables in Time Series Forecasting:}
MLP-based architectures like TiDE \cite{das2023long} and NBEATSx \cite{olivares2023neural} incorporate historical time series and exogenous variables to enhance forecasting accuracy. Classical methods such as ARIMAX \cite{vagropoulos2016comparison} and SARIMAX \cite{williams2001multivariate} also benefit from external inputs, with SARIMAX capturing seasonal patterns. Both of these models extend ARIMA by modeling complex dependencies via autoregressive, integrated, and moving average components under the influence of external factors. In the domain of time-series forecasting, a key challenge has been the effective integration of exogenous and endogenous variables. Traditional approaches often struggle with uneven sampling and missing values, primarily due to the direct concatenation of exogenous and endogenous inputs. TimeXer~\cite{wang2024timexer} addresses this issue by aggregating exogenous data into patch-wise endogenous representations, enabling the model to adapt to time-lagged and missing data more effectively. In contrast, our method applies a temporal window smoothing block to the exogenous inputs, producing a whitened representation based on globally significant patterns before integration with the endogenous variables.

\section{Methodology}

\noindent\textbf{Problem formulation:}
In the channel-independent approach \cite{PatchTST} for multivariate time series forecasting, historical observations, consisting of \( C \) distinct one-dimensional variables over \( L \) time steps, are represented as \( \mathbf{X} = \{\mathbf{x}_1, \dots, \mathbf{x}_L\} \in \mathbb{R}^{C \times L} \), where \( \mathbf{X} \) represent the matrix containing historical data for all features, and the \( i \)-th series can be represented as \( \mathbf{X}^{(i)} \in \mathbb{R}^{1 \times L} \). Furthermore, the model incorporates multiple exogenous series, denoted as \( \mathbf{E} = \{\mathbf{e}^1_{1:L_{ex}}, \mathbf{e}^2_{1:L_{ex}}, \dots, \mathbf{e}^N_{1:L_{ex}}\} \in \mathbb{R}^{N \times L_{\text{ex}}} \), where \( L_{\text{ex}} \) represents the look-back window length for the exogenous series. Here, \( \mathbf{E} \) comprises \( N \) independent exogenous variables, with each \( \mathbf{e}^i \in \mathbb{R}^{1 \times L_{ex}} \) corresponding to the \( i \)-th exogenous variable. The forecasting task aims to predict \( H \) future time steps of endogenous variables, represented as \(\widehat{\mathbf{\mathcal{Y}}} = \{\widehat{\mathbf{x}}_{L+1}, \dots, \widehat{\mathbf{x}}_{L+H}\} \in \mathbb{R}^{C \times H}\). To accomplish this, we seek to train a model $\mathcal{M}_\theta$, parameterized by $\theta$, that maps each series $\mathcal{S} = \{(\mathbf{X}, \mathbf{E})\}$ to the predicted horizon $\widehat{\mathbf{\mathcal{Y}}}$, such that:

\begin{equation}
\widehat{\mathbf{\mathcal{Y}}}_{L+1:L+H} = \mathcal{M}_\theta(\mathbf{X}_{1:L}, \mathbf{E}_{1:L_{\text{ex}}}).
\end{equation}

\subsection{Revisiting Models with exogenous input}
\begin{figure}[t]
  \centering
  \includegraphics[width=0.9\linewidth]{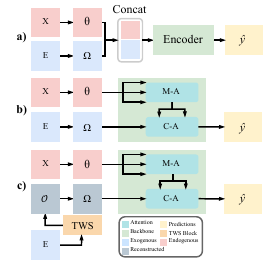} 
  \caption{Block diagram of (a-b) traditional approaches where raw exogenous input is directly integrated with endogenous input, in contrast to (c) our proposed method, which incorporates a smoothing process for exogenous input prior to its integration with endogenous input. 
  }
  \label{fig:revisiting}
\end{figure}
In several linear and deep learning models such as TiDE~\cite{das2023long} and NBEATSx~\cite{olivares2023neural}, exogenous inputs \( \mathbf{E} \) are first projected via a feature projection layer \( \theta(\mathbf{E}) \), then concatenated with endogenous variables \( \mathbf{X} \) to form \( \text{concat}(\mathbf{X}, \theta(\mathbf{E})) \), which is fed into the encoder, as illustrated in Fig.~\ref{fig:revisiting}(a). Transformer-based models such as TimeXer~\cite{wang2024timexer} process exogenous \( \mathbf{E} \) and endogenous \( \mathbf{X} \) inputs separately using distinct projection layers, \( \mathbf{\Omega}(\mathbf{E}) \) and \( \theta(\mathbf{X}) \), respectively. These representations are independently fed into the model \( \mathcal{M}_\theta(\theta(\mathbf{X}), \mathbf{\Omega}(\mathbf{E})) \), where multi-head attention (M-A) captures temporal dependencies and cross-attention (C-A) models variate-wise interactions, as shown in Fig.~\ref{fig:revisiting}(b). A global token then bridges both streams, enabling the exchange of causal information. However, none of these approaches refines or enhances the quality of exogenous variables, instead incorporates them in their raw form without improving their robustness. Unlike previous work, we preprocess the exogenous input, \(\mathbf{E}\) through our TWS block as shown in Fig. \ref{fig:revisiting}(c), to create a whitened version, \(\mathcal{O}\), based on key patterns that span the entire training set, making it more robust for forecasting. The refined exogenous, \( \mathcal{O} \), and the endogenous input, \( \mathbf{X} \), are then independently projected using separate feature projection layers, \( \theta(\mathbf{X}) \) and \( \mathbf{\Omega}(\mathcal{O}) \), respectively. These processed inputs are subsequently passed to the model, \( \mathcal{M}_\theta(\theta(\mathbf{X}),\mathbf{\Omega}(\mathcal{O})) \).

\begin{figure*}[hbt!]
  \centering
  \includegraphics[width=1\linewidth]{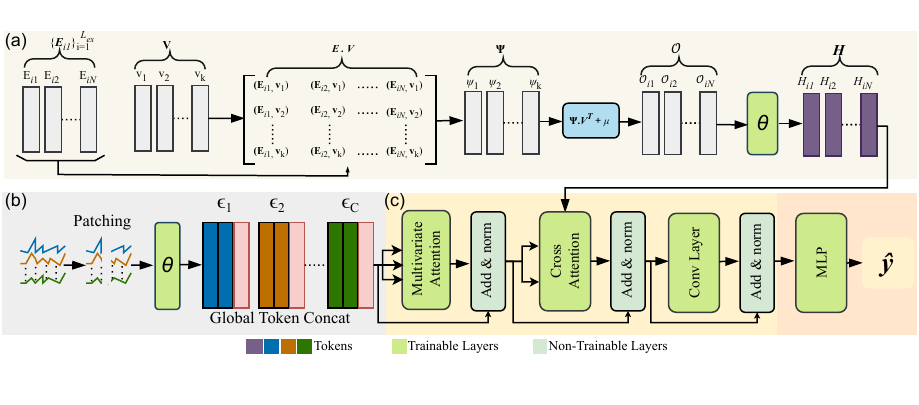}
  \caption{Overall architecture of the proposed method consists of three main components. \textbf{(a)} The exogenous series is whitened using our proposed method and projected into a latent space via a linear layer. \textbf{(b)} The endogenous series is split into patches, which are linearly projected to create patch embeddings, followed by concatenation of global tokens. \textbf{(c)} These representations are processed by a transformer-based architecture to to obtain the final predictions.}
  \label{fig:pretrain}
\end{figure*}

\subsection{Temporal Window Smoothing (TWS)}
Exogenous series for a specific window may contain redundant information and might not fully capture or reflect the global statistics of the training dataset. 
Based on this understanding, we propose a novel approach for exposing the exogenous series to a globally significant pattern to create a whitened version of it while mitigating redundant information that it may contain. This method, called TWS (Temporal Window Smoothing), transforms the original exogenous series for a specific window into a set of orthogonal components, each getting exposure to key patterns such as trends and seasonality that may span across the whole series. By leveraging these components, we reduce redundant information and make it more aware of global patterns, reconstructing a whitened, original-like exogenous series for each window to enhance its robustness for forecasting tasks.

\noindent\textbf{Exogenous Series Smoothing:}  
For smoothing the exogenous series for a specific window, we start with the entire training data \( \{\boldsymbol{X_{train}^i}\}_{i=1}^N \), where \( N \) is the number of features and \( \boldsymbol{train} \) refers to the training series. The data is mean-centered by subtracting the mean vector \( \boldsymbol{\mu} \) for each feature, calculated as: $ \mathbf{\tilde{X}}_{train} = \mathbf{X}_{train} - \boldsymbol{\mu}. $

To identify the key patterns in the whole series, we employ Principal Component Analysis (PCA), which finds a set of orthogonal basis that maximizes the explained variance in the data. We calculate the covariance matrix \( \Sigma \) of the centered data to capture the relationships between the variables throughout the series. It provides insight into how the features co-vary, helping to understand the structure and patterns within the data. The covariance matrix $\Sigma$ of the centered data is computed as: $ \boldsymbol{\Sigma} = \frac{1}{L_{seq}-1} \boldsymbol{\tilde{X}_{train}} \cdot \boldsymbol{\tilde{X}_{train}^{\top}} $, where \( L_{\text{seq}} \) denotes the length of the training sequence. Then we can express the eigendecomposition as:
$ \text{eig}(\boldsymbol{\Sigma}) = \{(\boldsymbol{\lambda}_1, \mathbf{v}_1), (\boldsymbol{\lambda}_2, \mathbf{v}_2), \ldots, (\boldsymbol{\lambda}_n, \mathbf{v}_n)\} $. where, $\mathbf{v}_1$ represents the principal eigenvector associated with the largest eigenvalue, $\mathbf{v}_2$ denotes the second eigenvector, and so forth. In addition, $\boldsymbol{\lambda}_1, \boldsymbol{\lambda}_2, \dots, \boldsymbol{\lambda}_n$ are the corresponding eigenvalues. The eigenvalues are ordered so that: $ \boldsymbol{\lambda}_1 \geq \boldsymbol{\lambda}_2 \geq \dots \geq \boldsymbol{\lambda}_N \geq 0 $.

Let \(\boldsymbol{U}\) denote the matrix of eigenvectors derived from the entire training set, encapsulating global statistical patterns. Our objective is to project the exogenous data into a subspace spanned by an \textit{optimal subset of column vectors} from \(\boldsymbol{U}\), selected based on their corresponding eigenvalues. These eigenvectors, computed over the complete training data, inherently represent key global patterns (e.g., trends, seasonality) and provide a set of orthogonal basis for representing statistically significant components. By projecting the exogenous series onto this subspace, we explicitly expose it to these globally important patterns, thereby aligning it with the intrinsic structure of the training data. 
This process involves discarding components that potentially contribute minimally to the global structure, retaining only the top-\( k \) components that capture the most significant patterns, such as trends and seasonality. However, the number of principal directions, \( \boldsymbol{k} \), to choose from the vector \( \boldsymbol{U} \) is a manually selected hyperparameter. This static selection may not necessarily yield the optimal result. To address this, the number of components, \( \boldsymbol{k} \), is dynamically determined for each dataset based on the cumulative explained variance ratio. This ensures that the selected subset of eigenvectors captures sufficient variance, aligning the exogenous series with the key global patterns while minimizing redundancy. Specifically, we select the smallest value of \( \boldsymbol{k} \) that retains at least 90\% of the variance such that: $\boldsymbol{V} = \frac{\sum_{i=1}^{j} \boldsymbol{\lambda}_i}{\sum_{i=1}^{N} \boldsymbol{\lambda}_i} \geq \boldsymbol{0.90}$.
Using \( \boldsymbol{V} = [\mathbf{v}_1, \mathbf{v}_2, \ldots, \mathbf{v}_k] \in \mathbb{R}^{N \times k} \), the exogenous input then projected onto a new subspace defined as:

\vspace{-0.5em}%
\begin{equation}
\boldsymbol{\Psi} = \boldsymbol{V}^{\top} \cdot \boldsymbol{E_{1:L_{ex}}},
\end{equation}%
\vspace{-0.5em}%
where each row \( \boldsymbol{\psi}_i \) of \( \boldsymbol{\Psi} \) represents a principal component:
\vspace{-0.1em}%
\[
\boldsymbol{\psi_i} = \mathbf{v_i^\top} \cdot \boldsymbol{E_{1:L_{ex}}}, \quad i = 1, \dots, k.
\]

The orthogonality of these components ensures minimal information overlap, while variance-based selection isolates globally significant patterns from less structured data.

\noindent\textbf{Reconstruction:}
By reconstructing the exogenous series using the optimal subset of components, we obtain a smoothened representation that is more aware of the essential global patterns while mitigating redundancy. Specifically, we project the data into a new subspace, retaining only the top-\( \mathbf{k} \) components that capture the most significant patterns. By utilizing these key components, a smoothened version of the exogenous series for each window, \( \mathcal{O} \), is obtained using the following equation: 
\begin{equation}
    \mathcal{O} = V\boldsymbol{\Psi} + \boldsymbol{\mu}
\end{equation}
where \( \mathcal{O} \in \mathbb{R}^{N \times L_{ex}} \). Adding the original mean vector \( \boldsymbol{\mu} \) ensures the reconstructed series is shifted back to the original scale. This approach leverages globally significant components, resulting in a robust representation that better captures underlying temporal dynamics and enhances forecasting performance.



\subsection{Proposed Architecture and Training}

\subsubsection{\textbf{Instance Normalization}} 
To mitigate training-test data distribution shifts, layer normalization~\cite{vaswani2017attention} is applied to both endogenous (\(X^{(i)}\)) and exogenous (\(E^{(i)}\)) inputs, standardizing them to zero mean and unit variance. During prediction, the original mean and variance are restored to preserve the data's original scale.

\subsubsection{\textbf{Endogenous Representation} }
For endogenous representation, we adopt a \textbf{channel-independent} approach \cite{PatchTST}, where each multivariate time series channel is processed independently to preserve its unique temporal characteristics.  Unlike \textbf{channel-mixing}, which aggregates data across channels, this method divides each time series \( \boldsymbol{X}^{(i)} \) into non-overlapping patches: $ \{ \boldsymbol{p}_1, \boldsymbol{p}_2, \dots, \boldsymbol{p}_{N_{en}} \} = \text{create\_patch}(\boldsymbol{X}) $ 

where \( \boldsymbol{p}_j \) is the \( j \)-th patch and \( N_{en} \) is the patch count. These patches are then mapped to a latent space via a linear projection layer:
\begin{equation}
    \boldsymbol{Z^i_{en}} = \text{LinearProjector}(\{ \boldsymbol{p}_1, \boldsymbol{p}_2, \dots, \boldsymbol{p}_{N_{en}} \}).
\end{equation}
A learnable global token \( \boldsymbol{\Phi}_{en} = \text{GlobalToken}(\mathbf{X}) \) is concatenated with the patches, acting as an intermediary between endogenous and exogenous series to convey causal information.

\subsubsection{\textbf{Exogenous Representation} }
Exogenous variables significantly enhance the forecasting performance of models by addressing irregularities such as missing values, misaligned timestamps, differing frequencies, and varying look-back lengths \cite{wang2024timexer}. However, traditional approaches to integrating exogenous variables with endogenous series might not be optimal. To overcome this limitation, we propose a \textbf{temporal smoothing block} that reduces redundancy in exogenous series while enhancing its awareness of global patterns. For each window \( \boldsymbol{E_{1:L_{ex}}} \), the refined representation is formulated as:
\begin{equation}
   \boldsymbol{\Psi} = \boldsymbol{E_{1:L_{ex}}} \times \mathbf{V}, \quad 
\boldsymbol{\mathcal{O}_{1:L_{ex}}} = \boldsymbol{\Psi} \mathbf{V}^T + \boldsymbol{\mu}. \label{eq:5} 
\end{equation}
Embedding for the refined exogenous series is generated as:
\begin{equation}
    \boldsymbol{H_{ex}} = \text{LinearProjector}(\boldsymbol{\mathcal{O}_{1:L_{ex}}})
\end{equation}
where, \( \boldsymbol{H_{ex}} = \{ \mathbf{h}_1, \dots, \mathbf{h}_N \} \in \mathbb{R}^{N \times D} \) represents \( N \) tokens embedded in a \( D \)-dimensional space. 

\subsubsection{\textbf{Self-Attention} }
The model employs self-attention to capture temporal dependencies within endogenous variable embeddings and utilizes a learnable global token as shown in Fig. \ref{fig:pretrain}(c) to integrate internal and external information. The global token operates in two phases: (a) it interacts with temporal tokens to gather patch-level information from the entire time series, and (b) each temporal token attends to the global token to receive information from the whitened exogenous input. The self-attention is formulated as:
\begin{equation}
    \boldsymbol{\hat{Z}^{(l)}_{\text{en}}}, \boldsymbol{\hat{\Phi}^{(l)}_{\text{en}}} = \text{Self-Attention}(\boldsymbol{[}\boldsymbol{Z^{(l)}_{\text{en}}}, \boldsymbol{\Phi^{(l)}_{\text{en}}}\boldsymbol{]})
\end{equation}

where \( \boldsymbol{[}\boldsymbol{Z^{(l)}_{\text{en}}}, \boldsymbol{\Phi^{(l)}_{\text{en}}}\boldsymbol{]} \) represents the concatenation of patch embeddings and global tokens along the sequence dimension, and \( l \in \{0,1,2, \dots, T-1\} \) denotes the \( l \)-th encoder block.

\begin{algorithm}[!t]
\caption{Proposed Time Series Prediction Method}
\label{algo:overall_architecture}
    \begin{algorithmic}[1]
    \Require 
        Endogenous input $\textbf{X} \in \mathbb{R}^{C \times L}$;
        Exogenous input $\textbf{E} \in \mathbb{R}^{N \times L_{ex}}$;
        Eigenvectors $\textbf{V} \in \mathbb{R}^{N \times k}$; Mean $\boldsymbol{\mu}$;
        Number of features $\textbf{N}$, $\textbf{C}$;
        Embedding dimension $\textbf{D}$;
        Transformer block number $\textbf{T}$; Number of patches $\boldsymbol{N_{en}}$.
        \State $\triangleright \ $ Split the endogenous series into $\boldsymbol{N_{en}}$ patches
        \State $\boldsymbol{\{ \boldsymbol{p}_1, \boldsymbol{p}_2, \dots, \boldsymbol{p}_{\boldsymbol{N}_{\boldsymbol{en}}} \}}
 = \text{create\_patch}(\mathbf{X})$  
        \State $\triangleright \ $ Create patch embeddings
        \State $\boldsymbol{Z^0_{en}} = \text{LinearProjector}(\boldsymbol{\{ \boldsymbol{p}_1, \boldsymbol{p}_2, \dots, \boldsymbol{p}_{\boldsymbol{N}_{\boldsymbol{en}}} \}}
)$
        \State $\triangleright \ $ Create a learnable global token for interaction with exogenous variables.
        \State $\boldsymbol{\Phi}_{en} = \text{GlobalToken}(\textbf{X})$
        \State $\triangleright \ $  Project the exogenous features in k dimensions 
        \State $ \boldsymbol{\Psi} = \boldsymbol{E_{1:L_{ex}}} \times \textbf{V} $
        \State $\triangleright \ $ Get exogenous features with reduced noise
        \State $ \boldsymbol{\mathcal{O}} = \boldsymbol{\Psi} \boldsymbol{V}^T + \boldsymbol{\mu}$
        \State $\triangleright \ $ Generate feature embeddings
        \State $ \boldsymbol{H^0} = \text{LinearProjector}(\boldsymbol{\mathcal{O}}) $
        \For {$\textbf{l} \gets 1$ to \textbf{T}} 
            \State $\boldsymbol{\hat{Z}^{(l-1)}_{\text{en}}}, \boldsymbol{\hat{\Phi}^{(l-1)}_{\text{en}}} = \text{Self-Attention}(\boldsymbol{[}\boldsymbol{Z^{(l-1)}_{\text{en}}}, \boldsymbol{\Phi^{(l-1)}_{\text{en}}}\boldsymbol{]})$
            \State $\boldsymbol{\hat\Phi^{(l-1)}_{\text{en}}} = \text{Cross-Attention}((\boldsymbol{\hat\Phi^{(l-1)}_{\text{en}}},\boldsymbol{H^{(l-1)}_{\text{ex}}))}$
            \State $\boldsymbol{Z^l_{\text{en}}}, \boldsymbol{\Phi^l_{\text{en}}} = \text{conv}(\boldsymbol{[}\boldsymbol{\hat{Z}^{(l-1)}_{\text{en}}}, \mathbf{\hat\Phi^{(l-1)}_{\text{en}}\boldsymbol{}])}$
        \EndFor
        
        \State $ \widehat{\mathbf{\mathcal{Y}}} = MLP(\boldsymbol{[}\boldsymbol{Z^{(T)}_{\text{en}}}, \mathbf{\Phi^{(T)}_{\text{en}}\boldsymbol{}])}$
        \State Return $\widehat{\mathbf{\mathcal{Y}}}$
    \end{algorithmic}
\end{algorithm}

\subsubsection{\textbf{Cross-Attention} }
The cross-attention mechanism uses the learned global token $\boldsymbol{\hat{\Phi}^{(l)}_{\text{en}}}$ from endogenous variables as queries and keys, while the exogenous variable embeddings $\boldsymbol{H^{(l)}_{ex}}$ as values, creating a bridge between these two sets of variables. The learned global token then aggregates information from exogenous variables. This can be formulated as: 
\begin{equation}
    \boldsymbol{\hat\Phi^{(l)}_{\text{en}}} = \text{Cross-Attention}((\boldsymbol{\hat\Phi^{(l)}_{\text{en}}},\boldsymbol{H^{(l)}_{\text{ex}}))}
\end{equation}

\subsubsection{\textbf{Conv-Layer}}
The learned global token \( \boldsymbol{\hat\Phi^{(l)}_{\text{en}}} \), obtained through cross-attention and with aggregated exogenous information, is further transformed by a convolutional layer along with the temporal tokens, expressed as:
\begin{equation}
    \boldsymbol{{Z}^{(l+1)}_{\text{en}}}, \boldsymbol{{\Phi}^{(l+1)}_{\text{en}}} = \text{conv}(\boldsymbol{[}\boldsymbol{\hat{Z}^{(l)}_{\text{en}}}, \boldsymbol{\hat\Phi^{(l)}_{\text{en}}}\boldsymbol{]})
\end{equation}
Finally, a flattening layer followed by a linear head is applied to generate the prediction result. which can be formulated as: 
\begin{equation}
 \widehat{\mathbf{\mathcal{Y}}} = MLP(\boldsymbol{[}\boldsymbol{Z^{(T)}_{\text{en}}}, \mathbf{\Phi^{(T)}_{\text{en}}\boldsymbol{}])}
\end{equation}

Here, $ \boldsymbol{[}\boldsymbol{Z^{(T)}_{\text{en}}}, \mathbf{\Phi^{(T)}_{\text{en}}\boldsymbol{}]} $ is the endogenous output embeddings from the final encoder layer.

\subsubsection{\textbf{Loss Function}}
To measure the difference between the predictions \( \widehat{\mathbf{\mathcal{Y}}} \) and the ground truth \( \mathbf{\mathcal{Y}} \), we employ Mean Squared Error (MSE). 
 The loss is computed for each channel, aggregated, and then averaged across \( M \) time series to determine the final objective loss. The loss can be calculated as: 
\begin{equation}
    \text{MSE} = \frac{1}{M} \sum_{i=1}^{M} (\widehat{\mathbf{\mathcal{Y}}}^{(i)}_{L+1:L+H} - {\mathbf{\mathcal{Y}}}^{(i)}_{L+1:L+H})^2
\end{equation}

Here, \( M \) represents the number series. \( {\mathbf{\mathcal{Y}}}^{(i)}_{L+1:L+H} \) and \( \widehat{\mathbf{\mathcal{Y}}}^{(i)}_{L+1:L+H} \) denote the ground truth and predicted values for the \( i \)-th series, respectively, on the forecasting horizon of length $\boldsymbol{H}$

\subsubsection{\textbf{Inference}}
During inference, we utilize the whitened version of exogenous variables. In this setup, only endogenous variables need to be predicted. The overall forecasting process can be expressed as:  
\begin{equation}
\widehat{\mathbf{\mathcal{Y}}}_{L+1:L+H} = \mathcal{M}_\theta(\mathbf{X}_{1:L}, TWS(\mathcal{O}_{1:L_{\text{ex}}})).
\end{equation}
Here, \( TWS \) refers to the exogenous input smoothing steps defined in Eqn. \ref{eq:5}.

\section{Experiments}
\subsection{Setup}
\noindent\textbf{Datasets:}
Our experiments are conducted on seven benchmark datasets: ECL, Traffic, ETT, and Weather. A detailed description of all datasets is provided in Table~\ref{tab:dataset_desc}


\textbf{ECL Dataset\footnote{\url{https://archive.ics.uci.edu/ml/datasets/ElectricityLoadDiagrams20112014/}}:}
{ The Electricity dataset contains hourly electricity consumption in kW per client from 2012 to 2014.}

\textbf{Traffic Dataset\footnote{\url{https://pems.dot.ca.gov/}}:}
{ The Traffic dataset details hourly road occupancy rates between 0 and 1 for 963 San Francisco Bay Area freeways, reflecting congestion levels. 
}

\textbf{ETT Dataset\footnote{\url{https://github.com/zhouhaoyi/ETDataset}}:}
{ The Electricity Transformer Temperature (ETT) dataset includes hour- and minute-level subsets to assess how different levels of data resolution impact Long Sequence Time-Series Forecasting.
\begin{itemize}
  \item Hour-level: Includes ETTh1 and ETTh2, which comprise hourly recorded data.
  \item Minute-level: Includes ETTm1 and ETTm2, featuring data sampled at 15-minute intervals.
\end{itemize}
}

\textbf{Weather Dataset\footnote{\url{https://www.bgc-jena.mpg.de/wetter/}}:}
{This dataset includes 21 meteorological variables, including air temperature and humidity, recorded every 10 minutes.}


\begin{table}[!t]
\centering
\setlength{\tabcolsep}{3.9pt} 
\renewcommand{\arraystretch}{1} 
\caption{Detailed dataset descriptions}
\begin{tabular}{lccc}
\toprule
Dataset        & \#Features & Dataset Size (Train / Test / Val) & Frequency \\
\midrule
ETTh1/ETTh2     & 7     & 8545 / 2881 / 2881                & Hourly \\
ETTm1/ETTm2     & 7     & 34465 / 11521 / 11521             & 15 minutes \\
Weather         & 21    & 36792 / 5271 / 10540              & 10 minutes \\
Traffic         & 862   & 12185 / 1757 / 3509              & Hourly \\
Electricity     & 321   & 18317 / 2633 / 5261              & Hourly \\
\bottomrule
\label{tab:dataset_desc}
\end{tabular}
\end{table}

\begin{table}[!t]
\centering
\renewcommand{\arraystretch}{0.6}
\caption{Ablation study on long-term forecasting task. Here, TWS represents our method. The best results are marked in \textbf{bold}}
\vspace{2pt}
\setlength{\tabcolsep}{2.5pt}
\resizebox{0.9\linewidth}{!}{
\begin{tabular}{c|c|c|cc|cc|cc}
\toprule[1.2pt]
\multirow{2}{*}{Bridging} & \multirow{2}{*}{TWS} 
& \multicolumn{2}{c}{ETTh1} & \multicolumn{2}{c}{ETTm1} & \multicolumn{2}{c}{Weather} \\
\cmidrule(lr){3-4}\cmidrule(lr){5-6}\cmidrule(lr){7-8}
&  & MSE & MAE & MSE & MAE & MSE & MAE \\ \toprule[1.2pt]
\multirow{2}{*}{{Concat}} &  w/o
& 0.439 & 0.434 & 0.387 & 0.399 & 0.255 & 0.277 \\ \cmidrule{2-8}
&  w
& 0.446 & 0.437 & 0.382 & 0.398 & 0.261 & 0.285 \\ \midrule

\multirow{2}{*}{Cross} &  w/o
& 0.445 & 0.437 & 0.381 & 0.396 & 0.248 & 0.276 \\ \cmidrule{2-8}
&  w
& \textbf{0.430} & \textbf{0.431} & \textbf{0.376} & \textbf{0.394} & \textbf{0.239} & \textbf{0.271} \\ \midrule

\end{tabular}}
\label{tab:abalation}
\vspace{-8pt}
\end{table}

\begin{table*}[htbp]
\tabcolsep=0.07cm
\centering
\caption{Multivariate long-term forecasting evaluated using MSE and MAE, where lower values indicate better results. The look-back length is fixed at 96, and the forecasting horizon is set to \( H \in \{96, 192, 336, 720\} \). The best results are highlighted in \textcolor{red}{red}, while the second-best results are highlighted in \textcolor{blue}{blue}}
\vspace{4pt}
\renewcommand{\multirowsetup}{\centering}
\setlength{\tabcolsep}{3pt}
\renewcommand{\arraystretch}{1.6}
\resizebox{\linewidth}{!}{
\begin{tabular}{cc|cc|cc|cc|cc|cc|cc|cc|cc|cc|cc|cc|cc|cc}
\toprule[1.2pt]  
\multicolumn{2}{c}{\multirow{2}{*}{\scalebox{1.25}{Models}}}
& \multicolumn{2}{c}{\scalebox{1.25}{Ours}}
& \multicolumn{2}{c}{\scalebox{1.25}{TimeXer \cite{wang2024timexer}}}
& \multicolumn{2}{c}{\scalebox{1.25}{iTrans. \cite{liu2023itransformer}}} 
& \multicolumn{2}{c}{\scalebox{1.25}{RLinear \cite{li2023revisiting}}} 
& \multicolumn{2}{c}{\scalebox{1.25}{PatchTST \cite{PatchTST}}} 
& \multicolumn{2}{c}{\scalebox{1.25}{Cross. \cite{zhang2022crossformer}}} 
& \multicolumn{2}{c}{\scalebox{1.25}{TiDE \cite{das2023long}}} 
& \multicolumn{2}{c}{\scalebox{1.25}{TimesNet \cite{Timesnet}}} 
& \multicolumn{2}{c}{\scalebox{1.25}{DLinear \cite{DLinear}}} 
& \multicolumn{2}{c}{\scalebox{1.25}{SCINet \cite{SCINet}}} 
& \multicolumn{2}{c}{\scalebox{1.25}{Stationary \cite{liu2023koopa}}} 
& \multicolumn{2}{c}{\scalebox{1.25}{Auto. \cite{Autoformer}}}  \\ 

\cmidrule(lr){3-4}
\cmidrule(lr){5-6}
\cmidrule(lr){7-8}
\cmidrule(lr){9-10}
\cmidrule(lr){11-12}
\cmidrule(lr){13-14}
\cmidrule(lr){15-16}
\cmidrule(lr){17-18}
\cmidrule(lr){19-20}
\cmidrule(lr){21-22}
\cmidrule(lr){23-24}
\cmidrule(lr){25-26}

\multicolumn{2}{c}{\scalebox{1.25}{Metric}}   
& \scalebox{1.25}{MSE}      & \scalebox{1.25}{MAE}
& \scalebox{1.25}{MSE}      & \scalebox{1.25}{MAE} 
& \scalebox{1.25}{MSE}      & \scalebox{1.25}{MAE}   
& \scalebox{1.25}{MSE}      & \scalebox{1.25}{MAE}    
& \scalebox{1.25}{MSE}      & \scalebox{1.25}{MAE} 
& \scalebox{1.25}{MSE}      & \scalebox{1.25}{MAE} 
& \scalebox{1.25}{MSE}      & \scalebox{1.25}{MAE}  
& \scalebox{1.25}{MSE}      & \scalebox{1.25}{MAE}      
& \scalebox{1.25}{MSE}      & \scalebox{1.25}{MAE}  
& \scalebox{1.25}{MSE}      & \scalebox{1.25}{MAE}
& \scalebox{1.25}{MSE}      & \scalebox{1.25}{MAE}     
& \scalebox{1.25}{MSE}      & \scalebox{1.25}{MAE}   \\ 

\toprule[1.2pt]
\multirow{5}{*}{\rotatebox{90}{\scalebox{1.25}{ECL}}} 
& \scalebox{1.25}{96}
& \scalebox{1.25}{\color{blue} 0.148} & \scalebox{1.25}{0.251}
& \scalebox{1.25}{\color{red} \textbf{0.140}} & \scalebox{1.25}{\color{blue} 0.242} 
 & \scalebox{1.25}{\color{blue} 0.148} & \scalebox{1.25}{\color{red} \textbf{0.240}} & 
 \scalebox{1.25}{0.201} & \scalebox{1.25}{0.281} & \scalebox{1.25}{0.195} & \scalebox{1.25}{0.285} & \scalebox{1.25}{0.219} & \scalebox{1.25}{0.314} & \scalebox{1.25}{0.237} & \scalebox{1.25}{0.329}  & \scalebox{1.25}{0.168} & \scalebox{1.25}{0.272} & \scalebox{1.25}{0.197} & \scalebox{1.25}{0.282} & \scalebox{1.25}{0.247} & \scalebox{1.25}{0.345} & \scalebox{1.25}{0.169} & \scalebox{1.25}{0.273} & \scalebox{1.25}{0.201} & \scalebox{1.25}{0.317} \\ 

& \scalebox{1.25}{192}
& \scalebox{1.25}{0.167} & \scalebox{1.25}{0.267}
& \scalebox{1.25}{\color{red} \textbf{0.157}} & \scalebox{1.25}{\color{blue} 0.256}
& \scalebox{1.25}{\color{blue} 0.162} & \scalebox{1.25}{\color{red} \textbf{0.253}} & 
\scalebox{1.25}{0.201} & \scalebox{1.25}{0.283} & \scalebox{1.25}{0.199} & \scalebox{1.25}{0.289} & \scalebox{1.25}{0.231} & \scalebox{1.25}{0.322} & \scalebox{1.25}{0.236} & \scalebox{1.25}{0.330} & \scalebox{1.25}{0.184} & \scalebox{1.25}{0.289} & \scalebox{1.25}{0.196} & \scalebox{1.25}{0.285} & \scalebox{1.25}{0.257} & \scalebox{1.25}{0.355} & \scalebox{1.25}{0.182} & \scalebox{1.25}{0.286} & \scalebox{1.25}{0.222} & \scalebox{1.25}{0.334} \\ 

& \scalebox{1.25}{336} 
& \scalebox{1.25}{0.183} & \scalebox{1.25}{0.283}
& \scalebox{1.25}{\color{red} \textbf{0.176}} & \scalebox{1.25}{\color{blue} 0.275}
& \scalebox{1.25}{\color{blue} 0.178} & \scalebox{1.25}{\color{red} \textbf{0.269}} & \scalebox{1.25}{0.215} & \scalebox{1.25}{0.298} & \scalebox{1.25}{0.215} & \scalebox{1.25}{0.305} & \scalebox{1.25}{0.246} & \scalebox{1.25}{0.337} & \scalebox{1.25}{0.249} & \scalebox{1.25}{0.344} & \scalebox{1.25}{0.198} & \scalebox{1.25}{0.300} & \scalebox{1.25}{0.209} & \scalebox{1.25}{0.301} & \scalebox{1.25}{0.269} & \scalebox{1.25}{0.369} & \scalebox{1.25}{0.200} & \scalebox{1.25}{0.304} & \scalebox{1.25}{0.231} & \scalebox{1.25}{0.338} \\ 

& \scalebox{1.25}{720} 
& \scalebox{1.25}{0.222} & \scalebox{1.25}{\color{blue} 0.316}
& \scalebox{1.25}{\color{red} \textbf{0.211}} & \scalebox{1.25}{\color{red} \textbf{0.306}}
& \scalebox{1.25}{0.225} & \scalebox{1.25}{0.317} & \scalebox{1.25}{0.257} & \scalebox{1.25}{0.331} & \scalebox{1.25}{0.256} & \scalebox{1.25}{0.337} & \scalebox{1.25}{0.280} & \scalebox{1.25}{0.363} & \scalebox{1.25}{0.284} & \scalebox{1.25}{0.373} & \scalebox{1.25}{\color{blue}0.220} & \scalebox{1.25}{0.320} & \scalebox{1.25}{0.245} & \scalebox{1.25}{0.333} & \scalebox{1.25}{0.299} & \scalebox{1.25}{0.390} & \scalebox{1.25}{0.222} & \scalebox{1.25}{0.321} & \scalebox{1.25}{0.254} & \scalebox{1.25}{0.361} \\

\cmidrule(lr){2-26}
& \scalebox{1.25}{Avg} 
& \scalebox{1.25}{0.180} & \scalebox{1.25}{\color{blue} 0.279}
& \scalebox{1.25}{\color{red} \textbf{0.171}} & \scalebox{1.25}{\color{red} \textbf{0.270}}
& \scalebox{1.25}{\color{blue} 0.178} & \scalebox{1.25}{\color{red} \textbf{0.270}} & \scalebox{1.25}{0.219} & \scalebox{1.25}{0.298} & \scalebox{1.25}{0.216} & \scalebox{1.25}{0.304} & \scalebox{1.25}{0.244} & \scalebox{1.25}{0.334} & \scalebox{1.25}{0.251} & \scalebox{1.25}{0.344} & \scalebox{1.25}{0.192} & \scalebox{1.25}{0.295} & \scalebox{1.25}{0.212} & \scalebox{1.25}{0.300} & \scalebox{1.25}{0.268} & \scalebox{1.25}{0.365} & \scalebox{1.25}{0.193} & \scalebox{1.25}{0.296} & \scalebox{1.25}{0.227} & \scalebox{1.25}{0.338} \\

\midrule

\multirow{5}{*}{\rotatebox{90}{\scalebox{1.25}{Weather}}} 
& \scalebox{1.25}{96}
& \scalebox{1.25}{\color{red} \textbf{0.154}} & \scalebox{1.25}{\color{red} \textbf{0.202}}
 & \scalebox{1.25}{\color{blue} 0.157} & \scalebox{1.25}{\color{blue} 0.205} & 
 \scalebox{1.25}{0.174} & \scalebox{1.25}{0.214} & \scalebox{1.25}{0.192} & \scalebox{1.25}{0.232} & \scalebox{1.25}{0.177} & \scalebox{1.25}{0.218} & \scalebox{1.25}{0.158} & \scalebox{1.25}{0.230}  & \scalebox{1.25}{0.202} & \scalebox{1.25}{0.261} & \scalebox{1.25}{0.172} & \scalebox{1.25}{0.220} & \scalebox{1.25}{0.196} & \scalebox{1.25}{0.255} & \scalebox{1.25}{0.221} & \scalebox{1.25}{0.306} & \scalebox{1.25}{0.173} & \scalebox{1.25}{0.223} & \scalebox{1.25}{0.266} & \scalebox{1.25}{0.336} \\ 

& \scalebox{1.25}{192}
& \scalebox{1.25}{\color{red} \textbf{0.204}} & \scalebox{1.25}{\color{blue} 0.250}
& \scalebox{1.25}{\color{red} \textbf{0.204}} & \scalebox{1.25}{\color{red}{\textbf{0.247}}} & 
\scalebox{1.25}{0.221} & \scalebox{1.25}{0.254} & \scalebox{1.25}{0.240} & \scalebox{1.25}{0.271} & \scalebox{1.25}{0.225} & \scalebox{1.25}{0.259} & \scalebox{1.25}{\color{blue} 0.206} & \scalebox{1.25}{0.277} & \scalebox{1.25}{0.242} & \scalebox{1.25}{0.298} & \scalebox{1.25}{0.219} & \scalebox{1.25}{0.261} & \scalebox{1.25}{0.237} & \scalebox{1.25}{0.296} & \scalebox{1.25}{0.261} & \scalebox{1.25}{0.340} & \scalebox{1.25}{0.245} & \scalebox{1.25}{0.285} & \scalebox{1.25}{0.307} & \scalebox{1.25}{0.367} \\ 

& \scalebox{1.25}{336} 
& \scalebox{1.25}{\color{red} \textbf{0.261}} & \scalebox{1.25}{\color{red} \textbf{0.290}}
& \scalebox{1.25}{\color{red} \textbf{0.261}} & \scalebox{1.25}{\color{red} \textbf{0.290}} & \scalebox{1.25}{0.278} & \scalebox{1.25}{\color{blue} 0.296} & \scalebox{1.25}{0.292} & \scalebox{1.25}{0.307} & \scalebox{1.25}{0.278} & \scalebox{1.25}{0.297} & \scalebox{1.25}{\color{blue} 0.272} & \scalebox{1.25}{0.335} & \scalebox{1.25}{0.287} & \scalebox{1.25}{0.335} & \scalebox{1.25}{0.280} & \scalebox{1.25}{0.306} & \scalebox{1.25}{0.283} & \scalebox{1.25}{0.335} & \scalebox{1.25}{0.309} & \scalebox{1.25}{0.378} & \scalebox{1.25}{0.321} & \scalebox{1.25}{0.338} & \scalebox{1.25}{0.359} & \scalebox{1.25}{0.395} \\

& \scalebox{1.25}{720} 
& \scalebox{1.25}{\color{red} \textbf{0.338}} & \scalebox{1.25}{\color{red} \textbf{0.340}}
& \scalebox{1.25}{\color{blue} 0.340} & \scalebox{1.25}{\color{blue} 0.341} & \scalebox{1.25}{0.358} & \scalebox{1.25}{0.349} & \scalebox{1.25}{0.364} & \scalebox{1.25}{0.353} & \scalebox{1.25}{0.354} & \scalebox{1.25}{0.348} & \scalebox{1.25}{0.398} & \scalebox{1.25}{0.418} & \scalebox{1.25}{0.351} & \scalebox{1.25}{0.386} & \scalebox{1.25}{0.365} & \scalebox{1.25}{0.359} & \scalebox{1.25}{{0.345}} & \scalebox{1.25}{0.381} & \scalebox{1.25}{0.377} & \scalebox{1.25}{0.427} & \scalebox{1.25}{0.414} & \scalebox{1.25}{0.410} & \scalebox{1.25}{0.419} & \scalebox{1.25}{0.428} \\ 

\cmidrule(lr){2-26}
& \scalebox{1.25}{Avg} 
& \scalebox{1.25}{\color{red} \textbf{0.239}} & \scalebox{1.25}{\color{red} \textbf{0.271}}
& \scalebox{1.25}{\color{blue} 0.241} & \scalebox{1.25}{\color{red} \textbf{0.271}} & \scalebox{1.25}{0.258} & \scalebox{1.25}{\color{blue} 0.279} & \scalebox{1.25}{0.272} & \scalebox{1.25}{0.291} & \scalebox{1.25}{0.259} & \scalebox{1.25}{0.281} & \scalebox{1.25}{0.259} & \scalebox{1.25}{0.315} & \scalebox{1.25}{0.271} & \scalebox{1.25}{0.320} & \scalebox{1.25}{0.259} & \scalebox{1.25}{0.287} &\scalebox{1.25}{0.265} & \scalebox{1.25}{0.317} & \scalebox{1.25}{0.292} & \scalebox{1.25}{0.363} & \scalebox{1.25}{0.288} & \scalebox{1.25}{0.314} & \scalebox{1.25}{0.338} & \scalebox{1.25}{0.382} \\ 

\midrule

\multirow{5}{*}{\rotatebox{90}{\scalebox{1.25}{ETTh1}}}
& \scalebox{1.25}{96} 
& \scalebox{1.25}{\color{red} \textbf{0.379}} & \scalebox{1.25}{\color{red} \textbf{0.393}}
& \scalebox{1.25}{\color{blue} 0.382} & \scalebox{1.25}{0.403} & \scalebox{1.25}{0.386} & \scalebox{1.25}{0.405} & \scalebox{1.25}{0.386} & \scalebox{1.25}{\color{blue} 0.395} & \scalebox{1.25}{0.414} & \scalebox{1.25}{0.419} & \scalebox{1.25}{0.423} & \scalebox{1.25}{0.448} & \scalebox{1.25}{0.479} & \scalebox{1.25}{0.464}  & \scalebox{1.25}{0.384} & \scalebox{1.25}{0.402} & \scalebox{1.25}{0.386} & \scalebox{1.25}{0.400} & \scalebox{1.25}{0.654} & \scalebox{1.25}{0.599} & \scalebox{1.25}{0.513} & \scalebox{1.25}{0.491} & \scalebox{1.25}{0.449} & \scalebox{1.25}{0.459}  \\ 

& \scalebox{1.25}{192} 
& \scalebox{1.25}{\color{red} \textbf{0.426}} & \scalebox{1.25}{\color{red} \textbf{0.423}}
& \scalebox{1.25}{\color{blue} 0.429} & \scalebox{1.25}{0.435} & \scalebox{1.25}{0.441} & \scalebox{1.25}{0.436} & \scalebox{1.25}{0.437} & \scalebox{1.25}{\color{blue} 0.424} & \scalebox{1.25}{0.460} & \scalebox{1.25}{0.445} & \scalebox{1.25}{0.471} & \scalebox{1.25}{0.474} & \scalebox{1.25}{0.525} & \scalebox{1.25}{0.492} & \scalebox{1.25}{0.436} & \scalebox{1.25}{0.429} & \scalebox{1.25}{0.437} & \scalebox{1.25}{0.432} & \scalebox{1.25}{0.719} & \scalebox{1.25}{0.631} & \scalebox{1.25}{0.534} & \scalebox{1.25}{0.504} & \scalebox{1.25}{0.500} & \scalebox{1.25}{0.482} \\ 

& \scalebox{1.25}{336} 
& \scalebox{1.25}{\color{red} \textbf{0.457}} & \scalebox{1.25}{\color{red} \textbf{0.445}}
& \scalebox{1.25}{\color{blue} 0.468} & \scalebox{1.25}{{0.448}} & \scalebox{1.25}{0.487} & \scalebox{1.25}{0.458} & \scalebox{1.25}{0.479} & \scalebox{1.25}{\color{blue} 0.446} & \scalebox{1.25}{0.501} & \scalebox{1.25}{0.466} & \scalebox{1.25}{0.570} & \scalebox{1.25}{0.546} & \scalebox{1.25}{0.565} & \scalebox{1.25}{0.515} & \scalebox{1.25}{0.491} & \scalebox{1.25}{0.469} & \scalebox{1.25}{0.481}  & \scalebox{1.25}{0.459} & \scalebox{1.25}{0.778} & \scalebox{1.25}{0.659} & \scalebox{1.25}{0.588} & \scalebox{1.25}{0.535} & \scalebox{1.25}{0.521} & \scalebox{1.25}{0.496} \\ 

& \scalebox{1.25}{720}
& \scalebox{1.25}{\color{red} \textbf{0.448}} & \scalebox{1.25}{\color{blue} 0.462}
& \scalebox{1.25}{\color{blue} 0.469} & \scalebox{1.25}{\color{red} \textbf{0.461}} & \scalebox{1.25}{0.503} & {\scalebox{1.25}{0.491}} & \scalebox{1.25}{{0.481}} & \scalebox{1.25}{{0.470}} & \scalebox{1.25}{0.500} & \scalebox{1.25}{0.488} & \scalebox{1.25}{0.653} & \scalebox{1.25}{0.621} & \scalebox{1.25}{0.594} & \scalebox{1.25}{0.558} & \scalebox{1.25}{0.521} & \scalebox{1.25}{0.500} & \scalebox{1.25}{0.519} & \scalebox{1.25}{0.516} & \scalebox{1.25}{0.836} & \scalebox{1.25}{0.699} & \scalebox{1.25}{0.643} & \scalebox{1.25}{0.616} & \scalebox{1.25}{0.514}  & \scalebox{1.25}{0.512}  \\

\cmidrule(lr){2-26}

& \scalebox{1.25}{Avg} 
& \scalebox{1.25}{\color{red} \textbf{0.427}} & \scalebox{1.25}{\color{red} \textbf{0.431}}
& \scalebox{1.25}{\color{blue}0.437} & \scalebox{1.25}{0.437} & \scalebox{1.25}{0.454} & \scalebox{1.25}{0.447} & \scalebox{1.25}{0.446} & \scalebox{1.25}{\color{blue} 0.434} & \scalebox{1.25}{0.469} & \scalebox{1.25}{0.454} & \scalebox{1.25}{0.529} & \scalebox{1.25}{0.522} & \scalebox{1.25}{0.541} & \scalebox{1.25}{0.507} & \scalebox{1.25}{0.458} & \scalebox{1.25}{0.450} & \scalebox{1.25}{0.456}  & \scalebox{1.25}{0.452} & \scalebox{1.25}{0.747} & \scalebox{1.25}{0.647} & \scalebox{1.25}{0.570} & \scalebox{1.25}{0.537} & \scalebox{1.25}{0.496} & \scalebox{1.25}{0.487}  \\ 

\midrule

\multirow{5}{*}{\rotatebox{90}{\scalebox{1.25}{ETTh2}}}
& \scalebox{1.25}{96} 
& \scalebox{1.25}{\color{red} \textbf{0.284}} & \scalebox{1.25}{\color{red} \textbf{0.337}}
& \scalebox{1.25}{\color{blue} 0.286} & \scalebox{1.25}{\color{blue} 0.338} & \scalebox{1.25}{0.297} & \scalebox{1.25}{0.349} & \scalebox{1.25}{{0.288}} & \scalebox{1.25}{\color{blue} 0.338} & \scalebox{1.25}{0.302} & \scalebox{1.25}{0.348} & \scalebox{1.25}{0.745} & \scalebox{1.25}{0.584} & \scalebox{1.25}{0.400} & \scalebox{1.25}{0.440} & \scalebox{1.25}{0.340} & \scalebox{1.25}{0.374} & \scalebox{1.25}{0.333} & \scalebox{1.25}{0.387} & \scalebox{1.25}{0.707} & \scalebox{1.25}{0.621} & \scalebox{1.25}{0.476} & \scalebox{1.25}{0.458} & \scalebox{1.25}{0.346} & \scalebox{1.25}{0.388} \\ 

& \scalebox{1.25}{192} 
& \scalebox{1.25}{\color{red} \textbf{0.362}} & \scalebox{1.25}{\color{red} \textbf{0.389}}
& \scalebox{1.25}{\color{blue} 0.363} & \scalebox{1.25}{\color{red} \textbf{0.389}} & \scalebox{1.25}{0.380} & \scalebox{1.25}{0.400} & \scalebox{1.25}{0.374} & \scalebox{1.25}{\color{blue} 0.390} & \scalebox{1.25}{0.388} & \scalebox{1.25}{0.400} & \scalebox{1.25}{0.877} & \scalebox{1.25}{0.656} & \scalebox{1.25}{0.528} & \scalebox{1.25}{0.509} & \scalebox{1.25}{0.402} & \scalebox{1.25}{0.414} & \scalebox{1.25}{0.477} & \scalebox{1.25}{0.476} & \scalebox{1.25}{0.860} & \scalebox{1.25}{0.689} & \scalebox{1.25}{0.512} & \scalebox{1.25}{0.493} & \scalebox{1.25}{0.456} & \scalebox{1.25}{0.452} \\ 

& \scalebox{1.25}{336}
& \scalebox{1.25}{0.418} & \scalebox{1.25}{0.432}
& \scalebox{1.25}{\color{red} \textbf{0.414}} & \scalebox{1.25}{\color{red} \textbf{0.423}} & \scalebox{1.25}{0.428} & \scalebox{1.25}{0.432} & \scalebox{1.25}{{\color{blue} 0.415}} & \scalebox{1.25}{\color{blue} 0.426} & \scalebox{1.25}{0.426} & \scalebox{1.25}{0.433} & \scalebox{1.25}{1.043} & \scalebox{1.25}{0.731} & \scalebox{1.25}{0.643} & \scalebox{1.25}{0.571}  & \scalebox{1.25}{0.452} & \scalebox{1.25}{0.452} & \scalebox{1.25}{0.594} & \scalebox{1.25}{0.541} & \scalebox{1.25}{1.000} & \scalebox{1.25}{0.744} & \scalebox{1.25}{0.552} & \scalebox{1.25}{0.551} & \scalebox{1.25}{0.482} & \scalebox{1.25}{0.486}\\ 

& \scalebox{1.25}{720} 
& \scalebox{1.25}{0.422} & \scalebox{1.25}{0.442}
& \scalebox{1.25}{\color{red} \textbf{0.408}} & \scalebox{1.25}{\color{red} \textbf{0.432}} & \scalebox{1.25}{0.427} & \scalebox{1.25}{0.445} & \scalebox{1.25}{{\color{blue} 0.420}} & \scalebox{1.25}{{\color{blue} 0.440}} & \scalebox{1.25}{0.431} & \scalebox{1.25}{0.446} & \scalebox{1.25}{1.104} & \scalebox{1.25}{0.763} & \scalebox{1.25}{0.874} & \scalebox{1.25}{0.679} & \scalebox{1.25}{0.462} & \scalebox{1.25}{0.468} & \scalebox{1.25}{0.831} & \scalebox{1.25}{0.657} & \scalebox{1.25}{1.249} & \scalebox{1.25}{0.838} & \scalebox{1.25}{0.562} & \scalebox{1.25}{0.560} & \scalebox{1.25}{0.515} & \scalebox{1.25}{0.511} \\ 

\cmidrule(lr){2-26}

& \scalebox{1.25}{Avg} 
& \scalebox{1.25}{\color{blue} 0.372} & \scalebox{1.25}{0.400}
& \scalebox{1.25}{\color{red} \textbf{0.368}} & \scalebox{1.25}{\color{red} \textbf{0.397}} & \scalebox{1.25}{0.383} & \scalebox{1.25}{0.407} & \scalebox{1.25}{{0.374}} & \scalebox{1.25}{\color{blue} 0.398} & \scalebox{1.25}{0.387} & \scalebox{1.25}{0.407} & \scalebox{1.25}{0.942} & \scalebox{1.25}{0.684} & \scalebox{1.25}{0.611} & \scalebox{1.25}{0.550} & \scalebox{1.25}{0.414} & \scalebox{1.25}{0.427} & \scalebox{1.25}{0.559} & \scalebox{1.25}{0.515} & \scalebox{1.25}{0.954} & \scalebox{1.25}{0.723} & \scalebox{1.25}{0.526} & \scalebox{1.25}{0.516} & \scalebox{1.25}{0.450} & \scalebox{1.25}{0.459} \\ 

\midrule

\multirow{5}{*}{\rotatebox{90}{\scalebox{1.25}{ETTm1}}}
& \scalebox{1.25}{96} 
& \scalebox{1.25}{\color{red} \textbf{0.317}} & \scalebox{1.25}{\color{blue} 0.357}
& \scalebox{1.25}{\color{blue} 0.318} & \scalebox{1.25}{\color{red} \textbf{0.356}} & \scalebox{1.25}{0.334} & \scalebox{1.25}{0.368} & \scalebox{1.25}{0.355} & \scalebox{1.25}{0.376} & \scalebox{1.25}{0.329} & \scalebox{1.25}{0.367} & \scalebox{1.25}{0.404} & \scalebox{1.25}{0.426} & \scalebox{1.25}{0.364} & \scalebox{1.25}{0.387} & \scalebox{1.25}{0.338} & \scalebox{1.25}{0.375} & \scalebox{1.25}{0.345} & \scalebox{1.25}{0.372} & \scalebox{1.25}{0.418} & \scalebox{1.25}{0.438} & \scalebox{1.25}{0.386} & \scalebox{1.25}{0.398} & \scalebox{1.25}{0.505} & \scalebox{1.25}{0.475} \\

& \scalebox{1.25}{192} 
& \scalebox{1.25}{\color{red} \textbf{0.355}} & \scalebox{1.25}{\color{red} \textbf{0.379}}
& \scalebox{1.25}{\color{blue} 0.362} & \scalebox{1.25}{\color{blue} 0.383} & \scalebox{1.25}{0.387} & \scalebox{1.25}{0.391} & \scalebox{1.25}{0.391} & \scalebox{1.25}{0.392} & \scalebox{1.25}{0.367} & \scalebox{1.25}{0.385} & \scalebox{1.25}{0.450} & \scalebox{1.25}{0.451} & \scalebox{1.25}{0.398} & \scalebox{1.25}{0.404} & \scalebox{1.25}{0.374} & \scalebox{1.25}{0.387} & \scalebox{1.25}{0.380} & \scalebox{1.25}{0.389} & \scalebox{1.25}{0.426} & \scalebox{1.25}{0.441} & \scalebox{1.25}{0.459} & \scalebox{1.25}{0.444} & \scalebox{1.25}{0.553} & \scalebox{1.25}{0.496} \\ 

& \scalebox{1.25}{336} 
& \scalebox{1.25}{\color{red} \textbf{0.385}} & \scalebox{1.25}{\color{red} \textbf{0.400}}
& \scalebox{1.25}{\color{blue} 0.395} & \scalebox{1.25}{\color{blue} 0.407} & \scalebox{1.25}{0.426} & \scalebox{1.25}{0.420} & \scalebox{1.25}{0.424} & \scalebox{1.25}{0.415} & \scalebox{1.25}{0.399} & \scalebox{1.25}{0.410} & \scalebox{1.25}{0.532} & \scalebox{1.25}{0.515} & \scalebox{1.25}{0.428} & \scalebox{1.25}{0.425} & \scalebox{1.25}{0.410} & \scalebox{1.25}{0.411}  & \scalebox{1.25}{0.413} & \scalebox{1.25}{0.413} & \scalebox{1.25}{0.445} & \scalebox{1.25}{0.459} & \scalebox{1.25}{0.495} & \scalebox{1.25}{0.464} & \scalebox{1.25}{0.621} & \scalebox{1.25}{0.537} \\ 

& \scalebox{1.25}{720} 
& \scalebox{1.25}{\color{red} \textbf{0.445}} & \scalebox{1.25}{\color{red} \textbf{0.438}}
& \scalebox{1.25}{\color{blue} 0.452} & \scalebox{1.25}{{0.441}} & \scalebox{1.25}{0.491} & \scalebox{1.25}{0.459} & \scalebox{1.25}{0.487} & \scalebox{1.25}{0.450} & \scalebox{1.25}{0.454} & \scalebox{1.25}{\color{blue} 0.439} & \scalebox{1.25}{0.666} & \scalebox{1.25}{0.589} & \scalebox{1.25}{0.487} & \scalebox{1.25}{0.461} & \scalebox{1.25}{0.478} & \scalebox{1.25}{0.450} & \scalebox{1.25}{0.474} & \scalebox{1.25}{0.453} & \scalebox{1.25}{0.595} & \scalebox{1.25}{0.550} & \scalebox{1.25}{0.585} & \scalebox{1.25}{0.516} & \scalebox{1.25}{0.671} & \scalebox{1.25}{0.561} \\

\cmidrule(lr){2-26}

& \scalebox{1.25}{Avg} 
& \scalebox{1.25}{\color{red} \textbf{0.376}} & \scalebox{1.25}{\color{red} \textbf{0.394}}
& \scalebox{1.25}{\color{blue} 0.382} & \scalebox{1.25}{\color{blue} 0.397} & \scalebox{1.25}{0.407} & \scalebox{1.25}{0.410} & \scalebox{1.25}{0.414} & \scalebox{1.25}{0.407} & \scalebox{1.25}{0.387} & \scalebox{1.25}{0.400} & \scalebox{1.25}{0.513} & \scalebox{1.25}{0.496} & \scalebox{1.25}{0.419} & \scalebox{1.25}{0.419} & \scalebox{1.25}{0.400} & \scalebox{1.25}{0.406} & \scalebox{1.25}{0.403} & \scalebox{1.25}{0.407} & \scalebox{1.25}{0.485} & \scalebox{1.25}{0.481} & \scalebox{1.25}{0.481} & \scalebox{1.25}{0.456} & \scalebox{1.25}{0.588} & \scalebox{1.25}{0.517} \\ 

\midrule

\multirow{5}{*}{\rotatebox{90}{\scalebox{1.25}{ETTm2}}}
& \scalebox{1.25}{96} 
& \scalebox{1.25}{\color{red} \textbf{0.169}} & \scalebox{1.25}{\color{red} \textbf{0.254}}
& \scalebox{1.25}{\color{blue} 0.171} & \scalebox{1.25}{\color{blue} 0.256} & \scalebox{1.25}{0.180} & \scalebox{1.25}{0.264} & \scalebox{1.25}{0.182} & \scalebox{1.25}{0.265} & \scalebox{1.25}{0.175} & \scalebox{1.25}{0.259} & \scalebox{1.25}{0.287} & \scalebox{1.25}{0.366} & \scalebox{1.25}{0.207} & \scalebox{1.25}{0.305} & \scalebox{1.25}{0.187} & \scalebox{1.25}{0.267} & \scalebox{1.25}{0.193} & \scalebox{1.25}{0.292} & \scalebox{1.25}{0.286} & \scalebox{1.25}{0.377} & \scalebox{1.25}{0.192} & \scalebox{1.25}{0.274} & \scalebox{1.25}{0.255} & \scalebox{1.25}{0.339} \\ 

& \scalebox{1.25}{192}
& \scalebox{1.25}{\color{red} \textbf{0.234}} & \scalebox{1.25}{\color{red} \textbf{0.297}}
& \scalebox{1.25}{\color{blue} 0.237} & \scalebox{1.25}{\color{blue} 0.299} & \scalebox{1.25}{0.250} & \scalebox{1.25}{0.309} & \scalebox{1.25}{0.246} & \scalebox{1.25}{0.304} & \scalebox{1.25}{0.241} & \scalebox{1.25}{0.302} & \scalebox{1.25}{0.414} & \scalebox{1.25}{0.492} & \scalebox{1.25}{0.290} & \scalebox{1.25}{0.364} & \scalebox{1.25}{0.249} & \scalebox{1.25}{0.309} & \scalebox{1.25}{0.284} & \scalebox{1.25}{0.362} & \scalebox{1.25}{0.399} & \scalebox{1.25}{0.445} & \scalebox{1.25}{0.280} & \scalebox{1.25}{0.339} & \scalebox{1.25}{0.281} & \scalebox{1.25}{0.340} \\ 

& \scalebox{1.25}{336} 
& \scalebox{1.25}{\color{blue} 0.298} & \scalebox{1.25}{\color{blue} 0.339}
& \scalebox{1.25}{\color{red} \textbf{0.296}} & \scalebox{1.25}{\color{red} \textbf{0.338}} & \scalebox{1.25}{0.311} & \scalebox{1.25}{0.348} & \scalebox{1.25}{0.307} & \scalebox{1.25}{0.342} & \scalebox{1.25}{0.305} & \scalebox{1.25}{0.343} & \scalebox{1.25}{0.597} & \scalebox{1.25}{0.542} & \scalebox{1.25}{0.377} & \scalebox{1.25}{0.422} & \scalebox{1.25}{0.321} & \scalebox{1.25}{0.351} & \scalebox{1.25}{0.369} & \scalebox{1.25}{0.427} & \scalebox{1.25}{0.637} & \scalebox{1.25}{0.591} & \scalebox{1.25}{0.334} & \scalebox{1.25}{0.361} & \scalebox{1.25}{0.339} & \scalebox{1.25}{0.372} \\

& \scalebox{1.25}{720} 
& \scalebox{1.25}{\color{red} \textbf{0.392}} & \scalebox{1.25}{\color{red} \textbf{0.394}}
& \scalebox{1.25}{\color{red} \textbf{0.392}} & \scalebox{1.25}{\color{red} \textbf{0.394}} & \scalebox{1.25}{0.412} & \scalebox{1.25}{0.407} & \scalebox{1.25}{0.407} & \scalebox{1.25}{\color{blue} 0.398} & \scalebox{1.25}{\color{blue} 0.402} & \scalebox{1.25}{0.400} & \scalebox{1.25}{1.730} & \scalebox{1.25}{1.042} & \scalebox{1.25}{0.558} & \scalebox{1.25}{0.524} & \scalebox{1.25}{0.408} & \scalebox{1.25}{0.403} & \scalebox{1.25}{0.554} & \scalebox{1.25}{0.522} & \scalebox{1.25}{0.960} & \scalebox{1.25}{0.735} & \scalebox{1.25}{0.417} & \scalebox{1.25}{0.413} & \scalebox{1.25}{0.433} & \scalebox{1.25}{0.432} \\

\cmidrule(lr){2-26}

& \scalebox{1.25}{Avg} 
& \scalebox{1.25}{\color{red} \textbf{0.273}} & \scalebox{1.25}{\color{red} \textbf{0.321}}
& \scalebox{1.25}{\color{blue} 0.274} & \scalebox{1.25}{\color{blue} 0.322} & \scalebox{1.25}{0.288} & \scalebox{1.25}{0.332} & \scalebox{1.25}{0.286} & \scalebox{1.25}{0.327} & \scalebox{1.25}{0.281} & \scalebox{1.25}{0.326} & \scalebox{1.25}{0.757} & \scalebox{1.25}{0.610} & \scalebox{1.25}{0.358} & \scalebox{1.25}{0.404} & \scalebox{1.25}{0.291} & \scalebox{1.25}{0.333} & \scalebox{1.25}{0.350} & \scalebox{1.25}{0.401} & \scalebox{1.25}{0.571} & \scalebox{1.25}{0.537} & \scalebox{1.25}{0.306} & \scalebox{1.25}{0.347} & \scalebox{1.25}{0.327} & \scalebox{1.25}{0.371} \\ 

\midrule

\multirow{5}{*}{\rotatebox{90}{\scalebox{1.25}{Traffic}}} 
& \scalebox{1.25}{96}
& \scalebox{1.25}{0.459} & \scalebox{1.25}{0.303}
& \scalebox{1.25}{\color{blue} 0.428} & \scalebox{1.25}{\color{blue} 0.271}
& \scalebox{1.25}{\color{red} \textbf{0.395}} & \scalebox{1.25}{\color{red} \textbf{0.268}} 
 & \scalebox{1.25}{0.649} & \scalebox{1.25}{0.389} & 
 \scalebox{1.25}{0.462} & \scalebox{1.25}{0.295} & \scalebox{1.25}{0.522} & \scalebox{1.25}{0.290} & \scalebox{1.25}{0.805} & \scalebox{1.25}{0.493} & \scalebox{1.25}{0.593} & \scalebox{1.25}{0.321}  & \scalebox{1.25}{0.650} & \scalebox{1.25}{0.396} & \scalebox{1.25}{0.788} & \scalebox{1.25}{0.499} & \scalebox{1.25}{0.612} & \scalebox{1.25}{0.338} & \scalebox{1.25}{0.613} & \scalebox{1.25}{0.388} \\ 

& \scalebox{1.25}{192}
& \scalebox{1.25}{0.478} & \scalebox{1.25}{0.308}
& \scalebox{1.25}{\color{blue} 0.448} & \scalebox{1.25}{\color{blue} 0.282}
& \scalebox{1.25}{\color{red} \textbf{0.417}} & \scalebox{1.25}{\color{red} \textbf{0.276}}
& \scalebox{1.25}{0.601} & \scalebox{1.25}{0.366} & \scalebox{1.25}{0.466} & \scalebox{1.25}{0.296} & \scalebox{1.25}{0.530} & \scalebox{1.25}{0.293} & \scalebox{1.25}{0.756} & \scalebox{1.25}{0.474} & \scalebox{1.25}{0.617} & \scalebox{1.25}{0.336} & \scalebox{1.25}{0.598} & \scalebox{1.25}{0.370} & \scalebox{1.25}{0.789} & \scalebox{1.25}{0.505} & \scalebox{1.25}{0.613} & \scalebox{1.25}{0.340} &
\scalebox{1.25}{0.616} & \scalebox{1.25}{0.382} \\ 

& \scalebox{1.25}{336} 
& \scalebox{1.25}{0.499} & \scalebox{1.25}{0.316}
& \scalebox{1.25}{\color{blue} 0.473} & \scalebox{1.25}{\color{blue} 0.289}
& \scalebox{1.25}{\color{red} \textbf{0.433}} & \scalebox{1.25}{\color{red} \textbf{0.283}}
& \scalebox{1.25}{0.609} & \scalebox{1.25}{0.369} & \scalebox{1.25}{0.482} & \scalebox{1.25}{0.304} & \scalebox{1.25}{0.558} & \scalebox{1.25}{0.305} & \scalebox{1.25}{0.762} & \scalebox{1.25}{0.477} & \scalebox{1.25}{0.629} & \scalebox{1.25}{0.336} & \scalebox{1.25}{0.605} & \scalebox{1.25}{0.373} & \scalebox{1.25}{0.797} & \scalebox{1.25}{0.508} & \scalebox{1.25}{0.618} & \scalebox{1.25}{0.328} & \scalebox{1.25}{0.622} & \scalebox{1.25}{0.337} \\ 

& \scalebox{1.25}{720} 
& \scalebox{1.25}{0.534} & \scalebox{1.25}{0.333}
& \scalebox{1.25}{\color{blue} 0.516} & \scalebox{1.25}{\color{blue} 0.307}
& \scalebox{1.25}{\color{red} \textbf{0.467}} & \scalebox{1.25}{\color{red} \textbf{0.302}}
& \scalebox{1.25}{0.647} & \scalebox{1.25}{0.387} & \scalebox{1.25}{0.514} & \scalebox{1.25}{0.322} & \scalebox{1.25}{0.589} & \scalebox{1.25}{0.328} & \scalebox{1.25}{0.719} & \scalebox{1.25}{0.449} & \scalebox{1.25}{0.640} & \scalebox{1.25}{0.350} & \scalebox{1.25}{0.645} & \scalebox{1.25}{0.394} & \scalebox{1.25}{0.841} & \scalebox{1.25}{0.523} & \scalebox{1.25}{0.653} & \scalebox{1.25}{0.355} &
\scalebox{1.25}{0.660} & \scalebox{1.25}{0.408} \\ 

\cmidrule(lr){2-26}
& \scalebox{1.25}{Avg} 
& \scalebox{1.25}{0.492} & \scalebox{1.25}{0.315}
& \scalebox{1.25}{\color{blue} 0.466} & \scalebox{1.25}{\color{blue} 0.287}
& \scalebox{1.25}{\color{red} \textbf{0.428}} & \scalebox{1.25}{\color{red} \textbf{0.282}} & \scalebox{1.25}{0.626} & \scalebox{1.25}{0.378} & \scalebox{1.25}{0.481} & \scalebox{1.25}{0.304} & \scalebox{1.25}{0.550} & \scalebox{1.25}{0.304} & \scalebox{1.25}{0.760} & \scalebox{1.25}{0.473} & \scalebox{1.25}{0.620} & \scalebox{1.25}{0.336} & \scalebox{1.25}{0.625} & \scalebox{1.25}{0.383} & \scalebox{1.25}{0.804} & \scalebox{1.25}{0.509} & \scalebox{1.25}{0.624} & \scalebox{1.25}{0.340} & \scalebox{1.25}{0.628} & \scalebox{1.25}{0.379} \\

\midrule

\multicolumn{2}{c|}
{\scalebox{1.25}{{$1^{\text{st}}$ Count}}} 
& \scalebox{1.25}{\color{red} \textbf{21}} & \scalebox{1.25}{\color{red} \textbf{18}} 
& \scalebox{1.25}{\color{blue}\textbf{12}} & \scalebox{1.25}{\color{blue}\textbf{13}} 
& \scalebox{1.25}{5} & \scalebox{1.25}{9} 
& \scalebox{1.25}{0} & \scalebox{1.25}{0} 
& \scalebox{1.25}{0} & \scalebox{1.25}{0} 
& \scalebox{1.25}{0} & \scalebox{1.25}{0} 
& \scalebox{1.25}{0} & \scalebox{1.25}{0} 
& \scalebox{1.25}{0} & \scalebox{1.25}{0} 
& \scalebox{1.25}{0} & \scalebox{1.25}{0} 
& \scalebox{1.25}{0} & \scalebox{1.25}{0} 
& \scalebox{1.25}{0} & \scalebox{1.25}{0} 
& \scalebox{1.25}{0} & \scalebox{1.25}{0}
\\ \bottomrule[1.2pt]          
\end{tabular}}
\label{tab:full-log-multi}
\end{table*}

\noindent\textbf{Evaluation metric:}
Following existing literature, such as TimeXer \cite{wang2024timexer}, PatchTST \cite{PatchTST}, and iTransformer \cite{liu2023itransformer}, we evaluate the performance of our proposed model using two widely recognized metrics: Mean Squared Error (MSE) and Mean Absolute Error (MAE). These metrics are computed for each window and averaged over all the windows. Lower values of MSE and MAE indicate better model performance.



\noindent\textbf{Baselines:}
We have compared the performance of our model with nine state-of-the-art deeplearning forecasting models for the multivariate forecasting task. The models include Transformer-based methods such as TimeXer\cite{wang2024timexer}, iTransformer\cite{liu2023itransformer}, PatchTST\cite{PatchTST}, Crossformer\cite{zhang2022crossformer}, and Autoformer\cite{Autoformer}. Linear-based methods include DLinear\cite{DLinear}, TiDE\cite{das2023long}, and RLinear\cite{li2023revisiting}. Additionally, TimesNet\cite{Timesnet}, a model based on the Temporal Convolutional Network.

\noindent\textbf{Implementation Details\footnote{Code is available at \url{https://github.com/robot-bulls/TWINS-E}}:}
We have trained and tested our model on a single NVIDIA RTX 4090 24GB GPU and Pytorch has been used for implementing all the models. We trained our proposed model using L2 loss as the objective function and the Adam \cite{kingma2014adam} as the optimizer. The patch length for endogenous input is set to 16. To ensure a fair comparison and prevent data leakage, we use the same procedure outlined in \cite{Timesnet} to create training, validation, and testing datasets. We compared our model with all the other models for the long-sequence multivariate forecasting task. For the forecast configuration, the length of the look-back series is fixed at 96 for all the data sets, with the prediction lengths set to 96, 192, 336, and 720. The number of epochs is set to 10, with early stopping applied if no improvement is observed for 3 consecutive epochs.


\subsection{Main results}

As shown in Table \ref{tab:full-log-multi}, our model outperforms TimeXer \cite{wang2024timexer} in four out of seven datasets in various prediction lengths, which incorporates exogenous input during prediction. Furthermore, our approach surpasses the second-best baseline, iTransformer \cite{liu2023itransformer}, in six out of seven datasets and consistently outperforms all other baselines in all datasets. This performance achivement can be attributed to our model’s ability to effectively integrate exogenous inputs while mitigating redundancy and enhancing global awareness—capabilities that address key limitations in existing methods.  When evaluated across 35 experimental settings for prediction lengths, our model achieves the best MSE performance in 21 settings, outperforming other models. TimeXer ranks second by achieving best results in 12 settings, followed by iTransformer with 5 settings.
In terms of MAE, our model achieves the best results in 18 settings, with TimeXer and iTransformer achieving 13 and 8 settings, respectively. Our approach demonstrates notable performance improvements across multiple datasets. Specifically, it achieves an average MSE reduction of 2.29\% on the ETTh1 dataset and an average MSE reduction of 1.57\%  on the ETTm1 dataset compared to the best baseline across all experimental forecasting lengths. 
Although relative improvements in specific settings may appear modest, the consistent superiority of our approach across various experimental configurations underscores its robustness and effectiveness.

\subsection{Ablation studies}

We conduct an ablation study to evaluate our proposed method by testing different strategies for integrating endogenous and exogenous series. Results, averaged across all prediction lengths, are shown in Table~\ref{tab:abalation}. For the concatenation (Concat) strategy, we remove the cross-attention (Cross) mechanism and global token, which are key to our architecture. Here, exogenous variable embeddings are directly concatenated with endogenous patch embeddings. To measure the effectiveness of our approach, we compare long-term forecasting performance with (w) and without (w/o) its implementation. Similarly, we evaluate the Cross-attention bridging technique, where a global token is employed to pass causal information from exogenous to endogenous inputs, to further validate the effectiveness of our proposed method. The last row in Table~\ref{tab:abalation} shows the performance of our method. Detailed ablation results for all forecasting horizons are provided in \textbf{Appendix}
\section{Conclusion}
This paper presents TWS, a novel temporal smoothing technique aimed to whiten exogenous inputs by leveraging global statistics while simultaneously reducing data redundancy. Recent work \cite{wang2024timexer} has demonstrated that the incorporation of exogenous series can significantly improve the forecasting performance in transformer-based models. Our proposed method builds on this insight by further improving the model's ability to capture global patterns and reducing redundancy, thereby amplifying the benefits of exogenous data incorporation.
In future work, we plan to explore the integration of TWS into a wider range of time series models across diverse domains, including linear-based models, to evaluate its broader applicability. Additionally, it would be valuable to investigate the potential of our approach in other tasks, such as short-term forecasting, imputation, and classification. We acknowledge that these directions offer promising opportunities for future research and will be addressed in subsequent studies.


\bibliographystyle{IEEEtran}
\bibliography{ref}

\begin{thebibliography}{10}
\providecommand{\url}[1]{#1}
\csname url@samestyle\endcsname
\providecommand{\newblock}{\relax}
\providecommand{\bibinfo}[2]{#2}
\providecommand{\BIBentrySTDinterwordspacing}{\spaceskip=0pt\relax}
\providecommand{\BIBentryALTinterwordstretchfactor}{4}
\providecommand{\BIBentryALTinterwordspacing}{\spaceskip=\fontdimen2\font plus
\BIBentryALTinterwordstretchfactor\fontdimen3\font minus \fontdimen4\font\relax}
\providecommand{\BIBforeignlanguage}[2]{{%
\expandafter\ifx\csname l@#1\endcsname\relax
\typeout{** WARNING: IEEEtran.bst: No hyphenation pattern has been}%
\typeout{** loaded for the language `#1'. Using the pattern for}%
\typeout{** the default language instead.}%
\else
\language=\csname l@#1\endcsname
\fi
#2}}
\providecommand{\BIBdecl}{\relax}
\BIBdecl

\bibitem{wu2023interpretable}
H.~Wu, H.~Zhou, M.~Long, and J.~Wang, ``Interpretable weather forecasting for worldwide stations with a unified deep model,'' \emph{Nature Machine Intelligence}, 2023.

\bibitem{zhang2023skilful}
Y.~Zhang, M.~Long, K.~Chen, L.~Xing, R.~Jin, M.~I. Jordan, and J.~Wang, ``Skilful nowcasting of extreme precipitation with nowcastnet,'' \emph{Nature}, 2023.

\bibitem{lago2021forecasting}
J.~Lago, G.~Marcjasz, B.~De~Schutter, and R.~Weron, ``Forecasting day-ahead electricity prices: A review of state-of-the-art algorithms, best practices and an open-access benchmark,'' \emph{Applied Energy}, vol. 293, p. 116983, 2021.

\bibitem{weron2014electricity}
R.~Weron, ``Electricity price forecasting: A review of the state-of-the-art with a look into the future,'' \emph{International journal of forecasting}, 2014.

\bibitem{zeng2023financial}
Z.~Zeng, R.~Kaur, S.~Siddagangappa, S.~Rahimi, T.~Balch, and M.~Veloso, ``Financial time series forecasting using cnn and transformer,'' \emph{arXiv preprint arXiv:2304.04912}, 2023.

\bibitem{lv2014traffic}
Y.~Lv, Y.~Duan, W.~Kang, Z.~Li, and F.-Y. Wang, ``Traffic flow prediction with big data: A deep learning approach,'' \emph{IEEE Transactions on Intelligent Transportation Systems}, vol.~16, no.~2, pp. 865--873, 2014.

\bibitem{vagropoulos2016comparison}
S.~I. Vagropoulos, G.~Chouliaras, E.~G. Kardakos, C.~K. Simoglou, and A.~G. Bakirtzis, ``Comparison of sarimax, sarima, modified sarima and ann-based models for short-term pv generation forecasting,'' in \emph{ENERGYCON}, 2016.

\bibitem{wang2024timexer}
Y.~Wang, H.~Wu, J.~Dong, Y.~Liu, Y.~Qiu, H.~Zhang, J.~Wang, and M.~Long, ``Timexer: Empowering transformers for time series forecasting with exogenous variables,'' \emph{Advances in Neural Information Processing Systems}, 2024.

\bibitem{williams2001multivariate}
B.~M. Williams, ``Multivariate vehicular traffic flow prediction: evaluation of arimax modeling,'' \emph{Transportation Research Record}, 2001.

\bibitem{das2023long}
A.~Das, W.~Kong, A.~Leach, R.~Sen, and R.~Yu, ``Long-term forecasting with tide: Time-series dense encoder,'' \emph{arXiv preprint arXiv:2304.08424}, 2023.

\bibitem{liu2023itransformer}
Y.~Liu, T.~Hu, H.~Zhang, H.~Wu, S.~Wang, L.~Ma, and M.~Long, ``itransformer: Inverted transformers are effective for time series forecasting,'' \emph{arXiv preprint arXiv:2310.06625}, 2023.

\bibitem{zhou2021informer}
H.~Zhou, S.~Zhang, J.~Peng, S.~Zhang, J.~Li, H.~Xiong, and W.~Zhang, ``Informer: Beyond efficient transformer for long sequence time-series forecasting,'' in \emph{AAAI}, 2021.

\bibitem{li2019enhancing}
S.~Li, X.~Jin, Y.~Xuan, X.~Zhou, W.~Chen, Y.-X. Wang, and X.~Yan, ``Enhancing the locality and breaking the memory bottleneck of transformer on time series forecasting,'' in \emph{NeurIPS}, 2019.

\bibitem{Timesnet}
H.~Wu, T.~Hu, Y.~Liu, H.~Zhou, J.~Wang, and M.~Long, ``Timesnet: Temporal 2d-variation modeling for general time series analysis,'' \emph{ICLR}, 2023.

\bibitem{devlin2018bert}
J.~Devlin, M.-W. Chang, K.~Lee, and K.~Toutanova, ``Bert: Pre-training of deep bidirectional transformers for language understanding,'' \emph{arXiv preprint arXiv:1810.04805}, 2018.

\bibitem{dosovitskiy2020image}
A.~Dosovitskiy, ``An image is worth 16x16 words: Transformers for image recognition at scale,'' \emph{arXiv preprint arXiv:2010.11929}, 2020.

\bibitem{liu2021swin}
Z.~Liu, Y.~Lin, Y.~Cao, H.~Hu, Y.~Wei, Z.~Zhang, S.~Lin, and B.~Guo, ``Swin transformer: Hierarchical vision transformer using shifted windows,'' in \emph{ICCV}, 2021.

\bibitem{Autoformer}
H.~Wu, J.~Xu, J.~Wang, and M.~Long, ``Autoformer: Decomposition transformers with {Auto-Correlation} for long-term series forecasting,'' \emph{NeurIPS}, 2021.

\bibitem{liu2021pyraformer}
S.~Liu, H.~Yu, C.~Liao, J.~Li, W.~Lin, A.~X. Liu, and S.~Dustdar, ``Pyraformer: Low-complexity pyramidal attention for long-range time series modeling and forecasting,'' in \emph{ICLR}, 2022.

\bibitem{zhou2022fedformer}
T.~Zhou, Z.~Ma, Q.~Wen, X.~Wang, L.~Sun, and R.~Jin, ``Fedformer: Frequency enhanced decomposed transformer for long-term series forecasting,'' in \emph{ICML}, 2022.

\bibitem{dong2023simmtm}
J.~Dong, H.~Wu, H.~Zhang, L.~Zhang, J.~Wang, and M.~Long, ``Simmtm: A simple pre-training framework for masked time-series modeling,'' in \emph{NeurIPS}, 2023.

\bibitem{PatchTST}
Y.~Nie, N.~H. Nguyen, P.~Sinthong, and J.~Kalagnanam, ``A time series is worth 64 words: Long-term forecasting with transformers,'' \emph{ICLR}, 2023.

\bibitem{olivares2023neural}
K.~G. Olivares, C.~Challu, G.~Marcjasz, R.~Weron, and A.~Dubrawski, ``Neural basis expansion analysis with exogenous variables: Forecasting electricity prices with nbeatsx,'' \emph{International Journal of Forecasting}, 2023.

\bibitem{vaswani2017attention}
A.~Vaswani, ``Attention is all you need,'' \emph{Advances in Neural Information Processing Systems}, 2017.

\bibitem{li2023revisiting}
Z.~Li, S.~Qi, Y.~Li, and Z.~Xu, ``Revisiting long-term time series forecasting: An investigation on linear mapping,'' \emph{arXiv preprint arXiv:2305.10721}, 2023.

\bibitem{zhang2022crossformer}
Y.~Zhang and J.~Yan, ``Crossformer: Transformer utilizing cross-dimension dependency for multivariate time series forecasting,'' in \emph{ICLR}, 2022.

\bibitem{DLinear}
A.~Zeng, M.~Chen, L.~Zhang, and Q.~Xu, ``Are transformers effective for time series forecasting?'' \emph{AAAI}, 2023.

\bibitem{SCINet}
M.~Liu, A.~Zeng, M.~Chen, Z.~Xu, Q.~Lai, L.~Ma, and Q.~Xu, ``Scinet: time series modeling and forecasting with sample convolution and interaction,'' \emph{NeurIPS}, 2022.

\bibitem{liu2023koopa}
Y.~Liu, C.~Li, J.~Wang, and M.~Long, ``Koopa: Learning non-stationary time series dynamics with koopman predictors,'' in \emph{NeurIPS}, 2023.

\bibitem{kingma2014adam}
D.~P. Kingma and J.~Ba, ``Adam: A method for stochastic optimization,'' \emph{arXiv preprint arXiv:1412.6980}, 2014.

\end{thebibliography}

\appendix

\section{Full Ablation Results}
This section provides the detailed results of the ablation study shown in Table. \ref{tab:abalation}. The best results are marked in \textcolor{red}{red} .
\begin{table}[ht]
\centering
\renewcommand{\arraystretch}{0.8}
\caption{Full Ablation study on long-term forecasting task.}
\vspace{2pt}
\setlength{\tabcolsep}{2.5pt}
\resizebox{0.9\linewidth}{!}{
\begin{tabular}{c|c|c|c|cc|cc|cc}
\toprule[1.2pt]
\multirow{2}{*}{Bridging} & \multirow{2}{*}{TWS} & \multirow{2}{*}{Horizon}
& \multicolumn{2}{c}{ETTh1} & \multicolumn{2}{c}{ETTm1} & \multicolumn{2}{c}{Weather} \\
\cmidrule(lr){4-5}\cmidrule(lr){6-7}\cmidrule(lr){8-9}
&  &  & MSE & MAE & MSE & MAE & MSE & MAE \\ \toprule[1.2pt]
\multirow{15}{*}{\rotatebox[origin=c]{90}{Concatenation}} &  \multirow{5}{*}{w/o} 
      & 96 & 0.378 & \textcolor{red}{0.393} & 0.337 & 0.367 & 0.159 & 0.206 \\
& & 192 & 0.430 & \textcolor{red}{0.423} & 0.362 & 0.383 & 0.222 & 0.256 \\
& & 336 & \textcolor{red}{0.454} & \textcolor{red}{0.442} & 0.394 & 0.405 & 0.279 & 0.297 \\
& & 720 & 0.493 & 0.480 & 0.456 & 0.440 & 0.358 & 0.348 \\ \cmidrule{2-9}
& & AVG & 0.439 & 0.434 & 0.387 & 0.399 & 0.255 & 0.277 \\ \cmidrule{2-9}

&  \multirow{5}{*}{w}
   & 96 & 0.379 & \textcolor{red}{0.393} & 0.328 & 0.365 & 0.158 & 0.206 \\
&  & 192 & 0.430 & 0.424 & 0.359 & 0.382 & 0.224 & 0.262 \\
&  & 336 & 0.464 & 0.444 & 0.393 & 0.404 & 0.282 & 0.304 \\
&  & 720 & 0.511 & 0.488 & 0.449 & 0.439 & 0.359 & 0.351 \\ \cmidrule{2-9}
&  & AVG & 0.446 & 0.437 & 0.382 & 0.398 & 0.261 & 0.285 \\ \midrule

\multirow{10}{*}{\rotatebox[origin=c]{90}{Cross-Attention}} &  \multirow{5}{*}{w/o}
   & 96 & \textcolor{red}{0.377} &	0.394 &	0.325 &	0.363 &	0.161 &	0.209 \\
&  & 192 & 0.427 &	\textcolor{red}{0.423} &	0.360 &	0.382 &	0.213 & 0.256 \\
&  & 336 & 0.486 &	0.455 & 0.390 &	0.402 &	0.268 &	0.294 \\
&  & 720 & 0.491 &	0.474 &	0.448 &	\textcolor{red}{0.438} &	0.348 &	0.345 \\ \cmidrule{2-9}
&  & AVG & 0.445 &	0.437 &	0.381 &	0.396 &	0.248 &	0.276 \\ \cmidrule{2-9}

&  \multirow{5}{*}{w}
          & 96 & 0.379 & \textcolor{red}{0.393} & \textcolor{red}{0.317} & \textcolor{red}{0.357} & \textcolor{red}{0.154} & \textcolor{red}{0.202} \\
&  & 192 & \textcolor{red}{0.426}  & \textcolor{red}{0.423} & \textcolor{red}{0.355} & \textcolor{red}{0.379} & \textcolor{red}{0.204} & \textcolor{red}{0.250} \\
&  & 336 & 0.457  & 0.445 & \textcolor{red}{0.385} & \textcolor{red}{0.400} & \textcolor{red}{0.261} & \textcolor{red}{0.290} \\
&  & 720 & \textcolor{red}{0.460}  & \textcolor{red}{0.465} & \textcolor{red}{0.445} & \textcolor{red}{0.438} & \textcolor{red}{0.338} & \textcolor{red}{0.340} \\ \cmidrule{2-9}
&  & AVG & \textcolor{red}{0.430} & \textcolor{red}{0.431} & \textcolor{red}{0.376} & \textcolor{red}{0.394} & \textcolor{red}{0.239} & \textcolor{red}{0.271} \\ \midrule
 
\end{tabular}}
\label{tab:fullabalation}
\vspace{-10pt}
\end{table}
\end{document}